\documentclass[lettersize,journal]{IEEEtran}
\usepackage{amsmath,amsfonts}
\usepackage{algorithmic}
\usepackage{algorithm}
\usepackage{array}
\usepackage[caption=false,font=normalsize,labelfont=sf,textfont=sf]{subfig}
\usepackage{textcomp}
\usepackage{stfloats}
\usepackage{url}
\usepackage{verbatim}
\usepackage{graphicx}
\usepackage{cite}
\usepackage{tikz,xcolor}
\begin{document}

\title{MotionVideoGAN: A Novel Video Generator Based on the Motion Space Learned from Image Pairs}

\author{Jingyuan Zhu, Huimin Ma, Jiansheng Chen, and Jian Yuan
\thanks{This work was supported in part by the National Natural Science Foundation of China under Grants U20B2062, 62172036, and 62227801 and in part by the National Key Research and Development Program of China under Grant 2022ZD0116300.  The associate editor coordinating the review of this manuscript and approving it for publication was Prof. Jingkuan Song. \emph{(Corresponding author: Huimin Ma.)}}
\thanks{J. Zhu, J. Yuan are with the Department of Electronic Engineering, Tsinghua University, Beijing 100084, China (e-mail: jy-zhu20@mails.tsinghua.edu.cn, jyuan@tsinghua.edu.cn).}
\thanks{H. Ma, J. Chen are with the School of Computer and Communication Engineering, University of Science and Technology Beijing, Beijing 100083, China (e-mail: mhmpub@ustb.edu.cn, jschen@ustb.edu.cn).}
\thanks{Digital Object Identifier 10.1109/TMM.2023.3251095}}

% \author{Jingyuan Zhu\orcidlink{0000-0003-3995-8382}, Huimin Ma\orcidlink{0000-0001-5383-5667}, Jiansheng Chen\orcidlink{0000-0002-2040-7938}, and Jian Yuan\orcidlink{0000-0001-9734-6056}
% \thanks{This work was supported in part by the National Natural Science Foundation of China under Grants U20B2062, 62172036, and 62227801 and in part by the National Key Research and Development Program of China under Grant 2022ZD0116300.  The associate editor coordinating the review of this manuscript and approving it for publication was Prof. Jingkuan Song. \emph{(Corresponding author: Huimin Ma.)}}
% \thanks{J. Zhu, J. Yuan are with the Department of Electronic Engineering, Tsinghua University, Beijing 100084, China (e-mail: jy-zhu20@mails.tsinghua.edu.cn, jyuan@tsinghua.edu.cn).}
% \thanks{H. Ma, J. Chen are with the School of Computer and Communication Engineering, University of Science and Technology Beijing, Beijing 100083, China (e-mail: mhmpub@ustb.edu.cn, jschen@ustb.edu.cn).}
% \thanks{Digital Object Identifier 10.1109/TMM.2023.3251095}}



\maketitle

\begin{abstract}
Video generation has achieved rapid progress benefiting from high-quality renderings provided by powerful image generators. We regard the video synthesis task as generating a sequence of images sharing the same contents but varying in motions. However, most previous video synthesis frameworks based on pre-trained image generators treat content and motion generation separately, leading to unrealistic generated videos. Therefore, we design a novel framework to build the motion space, aiming to achieve content consistency and fast convergence for video generation. We present MotionVideoGAN, a novel video generator synthesizing videos based on the motion space learned by pre-trained image pair generators. Firstly, we propose an image pair generator named MotionStyleGAN to generate image pairs sharing the same contents and producing various motions. Then we manage to acquire motion codes to edit one image in the generated image pairs and keep the other unchanged. The motion codes help us edit images within the motion space since the edited image shares the same contents with the other unchanged one in image pairs. Finally, we introduce a latent code generator to produce latent code sequences using motion codes for video generation. Our approach achieves state-of-the-art performance on the most complex video dataset ever used for unconditional video generation evaluation, UCF101. The source code is available on https://github.com/bbzhu-jy16/MotionVideoGAN.
\end{abstract}


\section{Introduction}
In recent years, image generation tasks have achieved significant success in generating high-quality \cite{Karras_2019_CVPR,Karras_2020_CVPR,Karras2021} and large-resolution \cite{karras2018progressive} images with wide variations in image contents \cite{pmlr-v97-zhang19d}. Modern image generation methods have been applied to related research fields, including image translation \cite{9721642,pang2021image,isola2017image,lee2018diverse,zhu2017unpaired,ma2018exemplar}, captioning \cite{9583266,9807935,8805998}, and novel view synthesis \cite{111,222,333,444,10.1145/3503927}. However, there still exists great room for improvement in video generation tasks. Unlike image generation tasks, video generation asks for a more complex target of generating a sequence of continuous images. That results in more complex networks, larger model sizes, and training costs. Early works \cite{Yushchenko_2019_ICCV,NIPS2016_04025959,Tulyakov_2018_CVPR,TGAN} attempt to generate temporally coherent videos from random noises with conv-based GAN \cite{NIPS2014_5ca3e9b1} models directly, leading to high computational costs and unsatisfying performance on large-resolution datasets. Video synthesis frameworks based on pre-trained image generators are proposed to pursue higher quality and larger resolution, including MoCoGAN-HD and StyleVideoGAN. They design motion generators to manipulate latent codes for synthesizing videos based on pre-trained image generators such as BigGAN \cite{DBLP:conf/iclr/BrockDS19} and StyleGAN2 \cite{Karras_2020_CVPR}. StyleGAN-V \cite{stylegan_v} proposes to generate videos by concatenating sequences of encoded motion codes to the constant input tensor of StyleGAN2. 

\begin{figure}[t]
\centering
\includegraphics[width=1.0\linewidth]{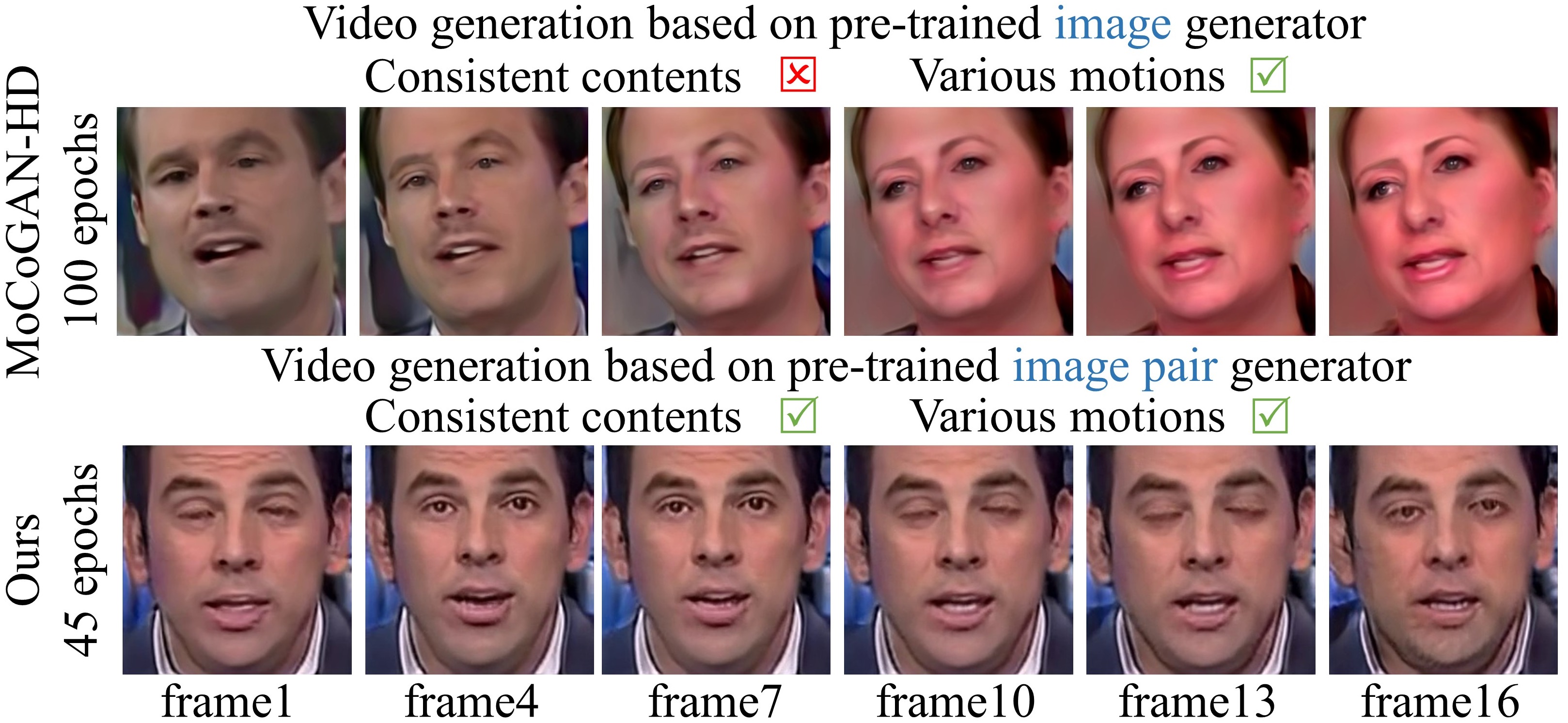}
\caption{Compared with MoCoGAN-HD based on pre-trained image generators, our approach achieves content consistency and fast convergence based on motions learned from image pairs.}
\label{motioncontent}
\end{figure}

As illustrated in MoCoGAN \cite{Tulyakov_2018_CVPR}, the latent space of GANs can be decomposed into content subspace and motion subspace. Following prior works, we focus on studying modern video generation datasets, where frames in a video share the same contents but vary in motions, e.g., persons talking in FaceForensics $256^2$ \cite{2018faceforensics} and cloud moving in SkyTimelapse $256^2$ \cite{Xiong_2018_CVPR}. Previous works \cite{tian2021a,stylegan_v} propose sequential generation methods to synthesize videos with image-based generators. Additional motion code generators or randomly sampled motion codes are needed for motion generation. They train generative models to fit the distribution of videos without unique methods of keeping contents consistent. The contents and motions are learned implicitly during training. We find that such methods result in inappropriate content editing when generating motions. In this paper, we propose a novel video generator using pre-trained image pair generators to build the motion space and naturally maintain content consistency in videos. As shown in the top row of Figure \ref{motioncontent}, MoCoGAN-HD \cite{tian2021a} produces videos containing inconsistent contents based on pre-trained image generators. More specifically, MoCoGAN-HD trained on FaceForensics $256^2$ aims to synthesize videos of a person speaking but makes the person change into another. Our approach naturally achieves content consistency and successfully produces high-quality videos of a single person talking when trained on FaceForensics $256^2$ with fewer training epochs as shown in the bottom row of Figure \ref{motioncontent}.

We build the proposed video generator, named MotionVideoGAN, as a three-stage approach for high-quality unconditional video generation. In the first stage, we sample image pairs from video datasets with fixed intervals and train the proposed image pair generator MotionStyleGAN on them to generate image pairs sharing the same contents and producing various motions. Secondly, we make use of the low-rank factorization \cite{zhu2021lowrankgan} of MotionStyleGAN's Jacobian matrix to obtain motion codes to edit one image in the generated image pairs and keep the other unchanged. Since two images in image pairs share the same contents, the editing is actually carried out in the motion space and keeps the contents unchanged. Finally, based on the motion codes, an LSTM-based \cite{6795963} latent code generator is employed to produce latent code sequences that are passed to MotionStyleGAN for video synthesis. Since the motion codes only manipulate one of the generated images within the motion space, image discriminators used in previous works \cite{Tulyakov_2018_CVPR,tian2021a} are no longer needed. The latent code generator is optimized with conv3d-based video discriminators to fit temporal evolution in real videos.

The main contributions of our work are summarized as follows: 

1) We propose the image pair generator MotionStyleGAN to learn contents and motions from image pairs simultaneously and generate image pairs sharing the same contents and producing various motions.

2) Our approach builds the motion space and naturally achieves content consistency for video generation using motion codes obtained through the low-rank factorization of MotionStyleGAN's Jacobian matrix.  

3) Based on the pre-trained MotionStyleGAN model and learned motion space, we present a novel video generator MotionVideoGAN with the advantages of content consistency and fast convergence, leading to superior improvement in complex scenes.

To adequately evaluate the effectiveness of our approach, we employ three modern video generation benchmarks: UCF101 $256^{2}$ \cite{soomro2012ucf101}, FaceForensics $256^{2}$ \cite{2018faceforensics}, and SkyTimelapse $256^2$ \cite{Xiong_2018_CVPR}. Frechet Video Distance (FVD) proposed by Thomas Unterthiner et al. \cite{unterthiner2019accurate} and Inception Score (IS) \cite{Saito_2017_ICCV} (only for UCF101 $256^2$) are used as the evaluation metrics for video generation. 

\section{Related Work}

\textbf{Video Synthesis}
Image generation tasks based on Generative Adversarial Nets (GANs) \cite{NIPS2014_5ca3e9b1} have achieved great success during the past few years. All kinds of network architectures \cite{RadfordMC15,NIPS2017_d0010a6f,Gong_2019_ICCV,2021you} and training methods \cite{NIPS2016_8a3363ab,NIPS2017_892c3b1c} have demonstrated GANs' ability to synthesize large-resolution \cite{karras2018progressive,10.1145/3343031.3350944} and high-quality \cite{DBLP:conf/iclr/BrockDS19,NEURIPS2019_18cdf49e,Karras_2019_CVPR,Karras_2020_CVPR,Karras2021} images. However, it remains a challenging problem to generate high-quality videos with large resolution and long video lengths unconditionally. Early works in the field of video generation, including VGAN \cite{NIPS2016_04025959} and DVDGAN \cite{DBLP:journals/corr/abs-1907-06571}, realize video generation with single-stage frameworks and achieve satisfying results on low-resolution datasets. TGAN \cite{Saito_2017_ICCV,TGAN} and MoCoGAN \cite{Tulyakov_2018_CVPR} decompose video generation into content generation and motion generation with content codes and motion codes generated from input noises. Many subsequent works \cite{acharya2018high,Yushchenko_2019_ICCV,Aich_2020_CVPR,Munoz_2021_WACV,tian2021a} follow similar strategies. SVGAN \cite{Sangeek_2021_CVPR} introduces self-supervision losses to improve video synthesis results. VideoGPT \cite{yan2021videogpt} makes use of VQ-VAE to learn downsampled discrete latent representations of raw videos and employs a GPT-like architecture to autoregressively model the discrete latent codes using spatio-temporal position encodings.

To further improve the quality of generated videos, MoCoGAN-HD \cite{tian2021a} and StyleVideoGAN \cite{fox2021stylevideogan} train motion generators for video generation based on pre-trained StyleGAN2 \cite{Karras_2020_CVPR} image generators. Both approaches deal with content and motion generation separately. MoCoGAN-HD \cite{tian2021a} calculates the principal components analysis (PCA) basis of randomly sampled motion codes in the latent space and trains an LSTM-based motion generator to obtain latent code sequences for video generation. StyleVideoGAN \cite{fox2021stylevideogan} employs Wasserstein GAN \cite{arjovsky2017wasserstein} for latent code sequence generation. The motions are learned implicitly during training. Unlike previous methods, our approach simultaneously learns contents and motions by training the MotionStyleGAN to generate image pairs and explicitly build the motion space for video generation. Based on the learned motion space, our approach naturally achieves content consistency in generated videos and realizes significant improvements in complex scenes.

StyleGAN-V \cite{stylegan_v} proposes an end-to-end video generation model built on top of StyleGAN2 with motion codes concatenated to its constant input tensor.  DIGAN \cite{yu2022generating} takes advantage of the implicit neural representations (INRs) and proposes a dynamics-aware implicit generative adversarial network. StyleGAN-V \cite{stylegan_v} and DIGAN \cite{yu2022generating} share similar ideas of exploring neural-based representations for continuous video synthesis. In addition, StyleGAN-V \cite{stylegan_v} replaces expensive conv3d blocks in video discriminators with a conv2d-based model, which sees only 2-4 randomly sampled frames per video to reduce computational costs and make it possible to generate larger resolution and longer videos. Other video generation methods propose approaches to simplify models and save computational costs as well. DVDGAN \cite{DBLP:journals/corr/abs-1907-06571} presents a computationally efficient decomposition of its discriminator. LDVDGAN \cite{KAHEMBWE2020506} employs a low-dimensional discriminator to reduce the model size and improve the resolution of generated videos. 

\begin{figure}[htbp]
\centering
\includegraphics[width=1.0\linewidth]{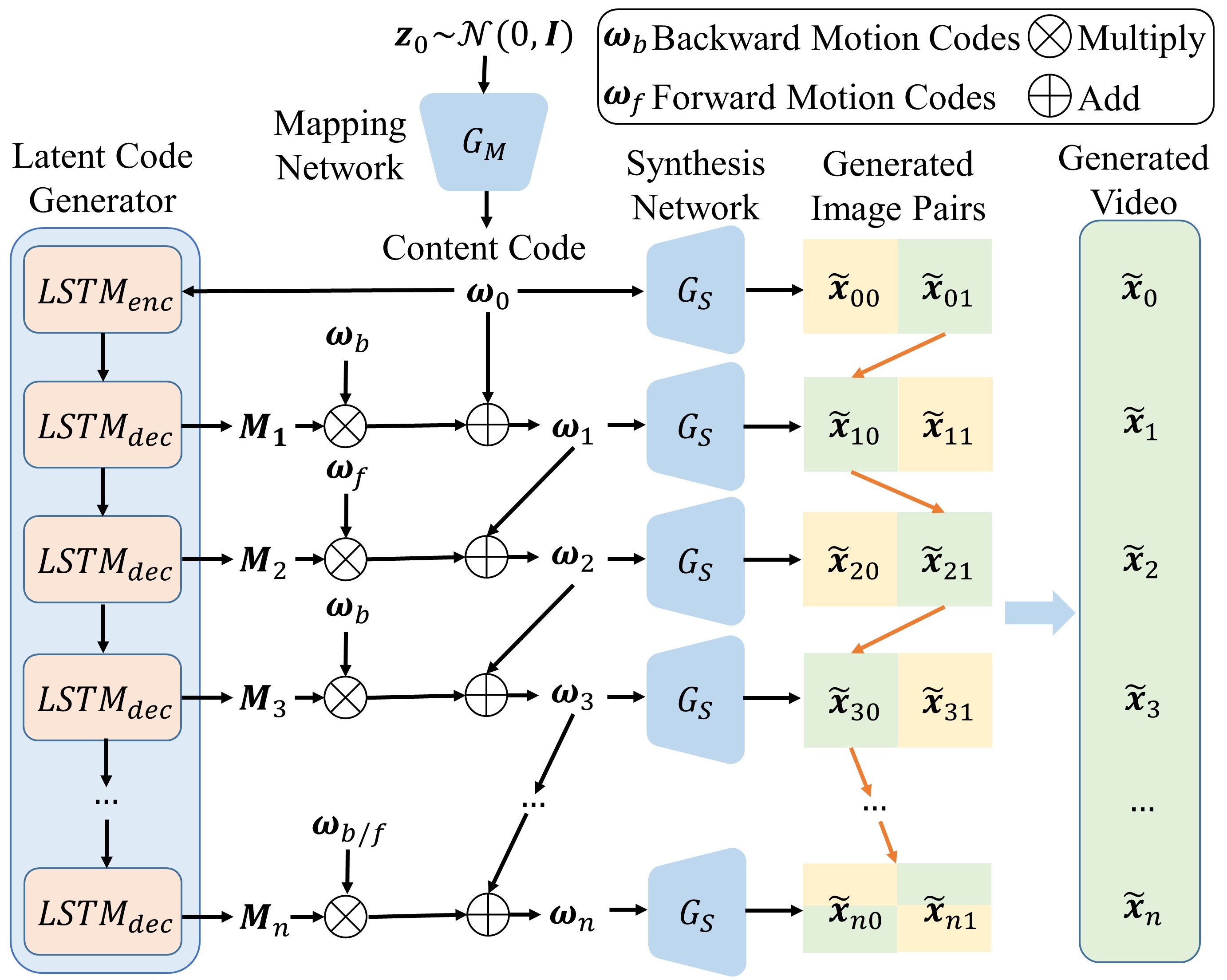}
\caption{MotionVideoGAN video generator architecture. $G_M$ and $G_S$ represent the mapping and synthesis network of the proposed MotionStyleGAN. $\boldsymbol{M}_t,t=1,2,...,n$ represent the coefficient matrices of motion codes $\boldsymbol{\omega}_b, \boldsymbol{\omega}_f$.  An LSTM-based latent code generator $L_{\omega}$ is employed to generate latent code sequences $\boldsymbol{\omega}_0,\boldsymbol{\omega}_1,...,\boldsymbol{\omega}_n$ based on the content codes and motion codes. The latent code sequences are passed to network $G_S$ for continuous video synthesis.}
\label{fig:temporal}
\end{figure}

\begin{figure}[htbp]
\centering
\includegraphics[width=1.0\linewidth]{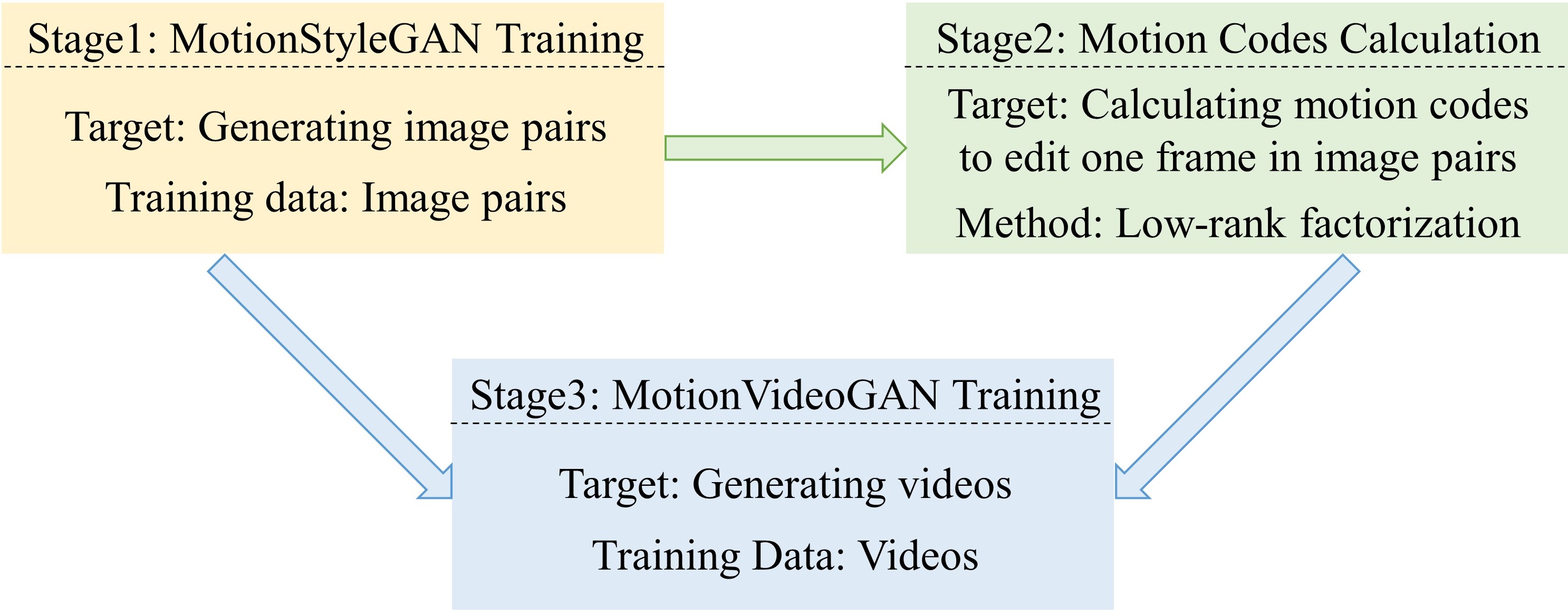}
\caption{Training pipeline of the proposed three-stage video generator.}
\label{fig:pipeline}
\end{figure}

\textbf{GAN-based Latent Space Embedding} 
Recently, interpreting well-trained GAN models is becoming widely concerned \cite{DBLP:conf/iclr/BauZSZTFT19}. Semantically meaningful vectors have been proven to exist in the latent space of GAN models \cite{pmlr-v119-voynov20a,Turkoglu_Thong_Spreeuwers_Kicanaoglu_2019,Goetschalckx_2019_ICCV,Karras_2019_CVPR,Shen_2020_CVPR,Jahanian*2020On,StyleSpace}. Besides, prior works successfully embed images back to the latent space of StyleGAN \cite{Abdal_2019_ICCV,Karras_2019_CVPR,Karras_2020_CVPR}.  Many works manipulate the generated results by editing the input latent codes \cite{Abdal_2020_CVPR,9241434,Plumerault2020Controlling,yu2021}. EditGAN \cite{ling2021editgan} can learn editing vectors in the latent space according to changes in semantic segmentations and generate corresponding changes in images with a GAN framework \cite{DatasetGAN,SemanticGAN} that jointly models images and their semantic segmentations. LowRankGAN \cite{zhu2021lowrankgan} is the first to demonstrate the effectiveness of low-rank factorization in interpreting GANs. They obtain attribute vectors to perform precise local editing in images while keeping other areas stable. Our method employs the low-rank factorization of MotionStyleGAN's Jacobian matrix and obtains motion codes for video generation following similar methods. 

\section{Method}
\label{method}

In this section, we introduce our video generator MotionVideoGAN in detail. More specifically, our framework can be divided into three stages. Firstly, the proposed image pair generator MotionStyleGAN $G_P$ built on top of StyleGAN2 \cite{Karras_2020_CVPR} is trained to learn image contents and reasonable motions from image pairs with fixed interval $k$ in video datasets. For an arbitrary input noise $\boldsymbol{z}_t$ ($t$ represents an arbitrary time step), we can generate image pairs as $\left[ \tilde{\boldsymbol{x}}_{t}, \tilde{\boldsymbol{x}}_{t\pm k} \right] = G_P(\boldsymbol{z}_t)$. The image pair generation can be divided into two steps:
\begin{equation}
    \begin{aligned}
        \boldsymbol{\omega}_t &= G_M(\boldsymbol{z}_t), \quad
        \left[ \tilde{\boldsymbol{x}}_{t}, \tilde{\boldsymbol{x}}_{t\pm k} \right]&=G_S(\boldsymbol{\omega}_t),
    \end{aligned}
\end{equation}
where $\boldsymbol{\omega}_t$ represents the corresponding latent code of $\boldsymbol{z}_t$, $G_M$ and $G_S$ represent the mapping network and synthesis network of MotionStyleGAN, respectively. Then we use the low-rank factorization of MotionStyleGAN's Jacobian matrix to calculate motion codes that can edit the former or the latter frame in image pairs and maintain the other unchanged. Finally, an LSTM-based latent code generator $L_{\omega}$ is employed for producing latent code sequences $\left\lbrace\boldsymbol{\omega}_{0},\boldsymbol{\omega}_{1},...,\boldsymbol{\omega}_{n}\right\rbrace$ based on content codes and motion codes, where $n$ is the number of frames in videos. Continuous videos are synthesized by $G_S$ with latent code sequences. Video discriminators composed of conv3d-based networks are applied to $L_{\omega}$ training, helping the generated videos fit temporal evolution in training videos. The video generator architecture and its three-stage training pipeline are illustrated in Figure \ref{fig:temporal} and \ref{fig:pipeline}, respectively. The following parts describe each stage of the proposed MotionVideoGAN in detail.

\subsection{MotionStyleGAN}
\label{motionstylegan}
\begin{figure}[t]
\centering
\includegraphics[width=.8\linewidth]{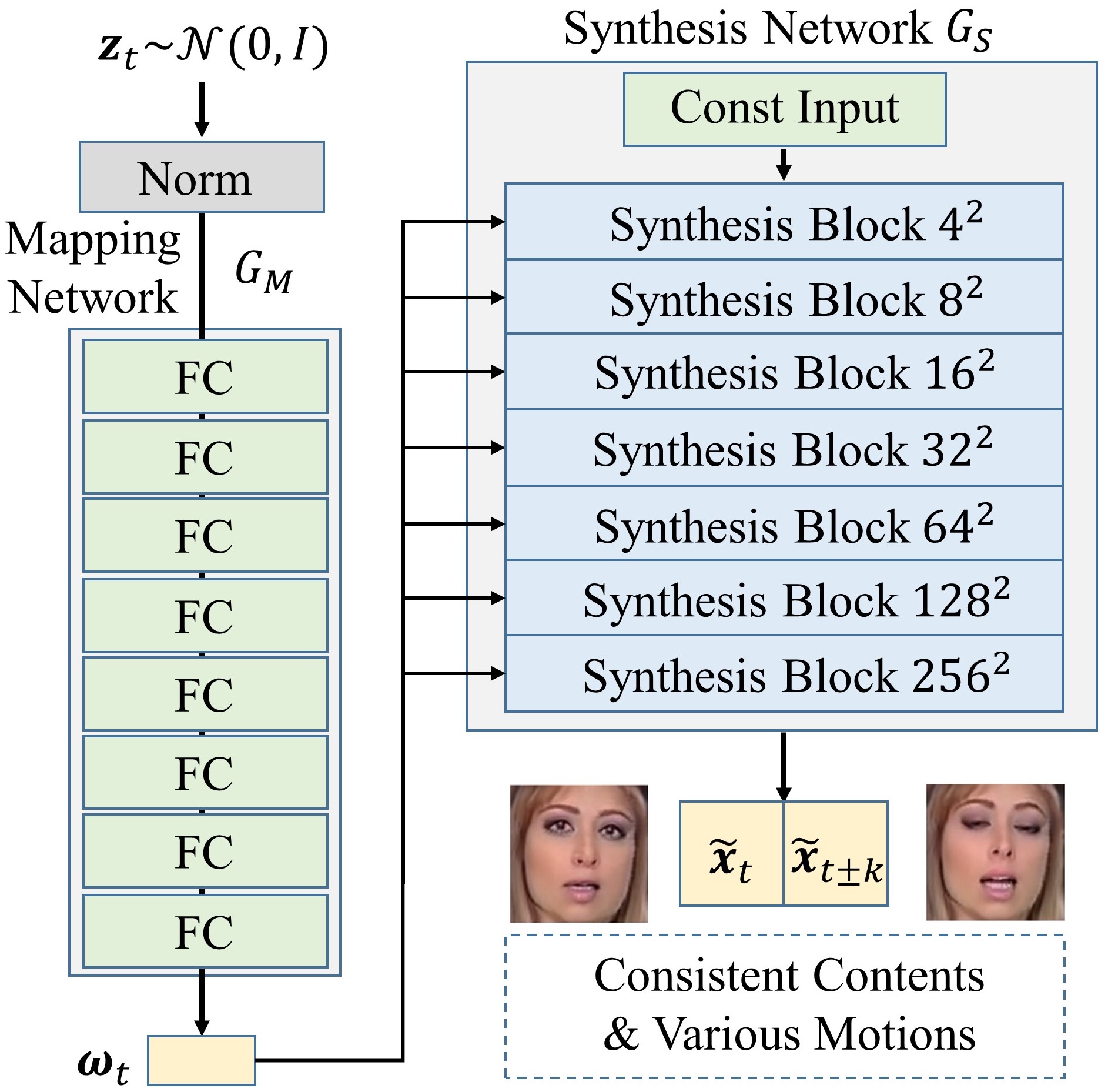}
\caption{MotionStyleGAN network architecture. The proposed model is built on top of StyleGAN2 \cite{Karras_2020_CVPR}. We modify StyleGAN2 to generate image pairs $\left[ \tilde{\boldsymbol{x}}_{t}, \tilde{\boldsymbol{x}}_{t\pm k} \right]$ sharing the same contents and producing various motions. The MotionStyleGAN consists of mapping network $G_M$ (multi-layer fully connected layers) and synthesis network $G_S$. It is trained with image pairs sampled from video datasets with fixed interval $k$. }
\label{fig:motionstylegan}
\end{figure}

The proposed image pair generator MotionStyleGAN is built on top of StyleGAN2. We modify the network architectures of StyleGAN2 to generate and discriminate six color channels for image pairs rather than three channels (RGB) for a single image. The generator architecture, including mapping network $G_M$ and synthesis network $G_S$, is shown in Figure \ref{fig:motionstylegan}. It is trained to generate image pairs sharing the same contents and producing various motions under the guidance of image pair discriminators. The training video datasets $\emph{D}=\left\lbrace \boldsymbol{x}_{t}\right\rbrace_{t=1}^{n}$ are reorganized as image pairs $\left[ \boldsymbol{x}_{t},\boldsymbol{x}_{t+k}\right]$ and $\left[ \boldsymbol{x}_{t+k},\boldsymbol{x}_{t} \right], 1\leq t,t+k\leq n$ with fixed interval $k$ for motion learning. With reorganized datasets, the MotionStyleGAN is trained to learn bidirectional motions. The intervals are set empirically, ensuring that typical motions like winking and talking in FaceForensics $256^2$ \cite{2018faceforensics} and cloud moving in SkyTimelapse $256^2$ \cite{Xiong_2018_CVPR} can be learned by MotionStyleGAN from the reorganized image pair datasets. 

Modern video generators \cite{tian2021a,fox2021stylevideogan} mainly depend on pre-trained image generators to improve video quality. Therefore, other methods are needed to keep content consistency and generate appropriate motions, leading to additional computational costs and unrealistic results. As for the proposed MotionStyleGAN, we train it to learn contents and motions simultaneously from image pairs with fixed intervals. Moreover, since two images in a pair of generated images share the same contents, we can keep the consistency of contents and generate motions by keeping one of them unchanged and editing the other.

\subsection{Motion Code}
\label{motioncode}
 Letting $\boldsymbol{\omega}_t=G_{M}(\boldsymbol{z}_t)$ denote the content code, we can generate image pairs as $\left[ \tilde{\boldsymbol{x}}_{t}, \tilde{\boldsymbol{x}}_{t\pm k} \right] = G_{S}(\boldsymbol{\omega}_t)$ with MotionStyleGAN. Prior works \cite{ling2021editgan,zhu2021lowrankgan} have proved it possible to realize local editing with manipulation in the latent space of GAN models. As for our approach, we manipulate the content code $\boldsymbol{\omega}_t \in \mathbb{R}^{d_{\omega}}$ in the latent space of MotionStyleGAN with motion codes to achieve motion generation and keep the contents unchanged. Since image pairs generated by the MotionStyleGAN share the same contents, editing one frame can only change its motions and keep the original contents. If we have:

 \begin{equation}
 \begin{aligned}
     \left[ \tilde{\boldsymbol{x}}_{t}, \tilde{\boldsymbol{x}}_{t\pm k}^{edit} \right] &= G_{S}(\boldsymbol{\omega}_t+\alpha \boldsymbol{\omega}_{f}), \\
     \left[ \tilde{\boldsymbol{x}}_{t}^{edit}, \tilde{\boldsymbol{x}}_{t\pm k} \right] &= G_{S}(\boldsymbol{\omega}_t+\alpha \boldsymbol{\omega}_{b}),
 \end{aligned}
 \end{equation}
 where $\boldsymbol{\omega}_{b}$ and $\boldsymbol{\omega}_{f}$ denote attribute vectors editing the former or latter frame and $\alpha$ represents the editing strength, the attribute vectors $\boldsymbol{\omega}_{b}$ and $\boldsymbol{\omega}_{f}$ are exactly the motion codes needed for motion generation. 

We employ the low-rank factorization \cite{zhu2021lowrankgan} of MotionStyleGAN's Jacobian matrix for motion code generation. Letting the real images $\boldsymbol{x} \in \mathbb{R}^{d_x}$ be in the $d_{x}$-dimensional space and the latent code $\boldsymbol{\omega}_t \in \mathbb{R}^{d_{\omega}}$ be of dimension $d_{\omega}$, the Jacobian matrix $\boldsymbol{J}_{\omega}$ of the synthesis network in MotionStyleGAN is defined as:
\begin{equation}
    (\boldsymbol{J}_{\omega})_{i,j}=\frac{\partial G_S(\boldsymbol{\omega}_t)_i}{\partial (\omega_t)_j} 
\end{equation}
for an arbitrary latent code $\boldsymbol{\omega}_t$, where $(\boldsymbol{J}_{\omega})_{i,j}$ represents the $(i,j)$-th entry of  $\boldsymbol{J}_{\omega} \in \mathbb{R}^{2d_x \times d_{\omega}}$. Then the first-order approximation to the result edited with an attribute vector $\boldsymbol{n}$  can be written as:
\begin{equation}
    G_S(\boldsymbol{\omega}_t+\alpha \boldsymbol{n})=G_S(\boldsymbol{\omega}_t)+\alpha \boldsymbol{J}_{\omega} \boldsymbol{n} + o(\alpha)
\end{equation}
by the Taylor series. By maximizing the following variance, effective editing directions $\boldsymbol{n}$ can be obtained.

\begin{equation}
\begin{aligned}
    || G_S(\boldsymbol{\omega}_t+ & \alpha \boldsymbol{n})  -G_S(\boldsymbol{\omega}_t) ||_2^{2} \approx \alpha^2 \boldsymbol{n}^{T}\boldsymbol{J}_{\omega}^T \boldsymbol{J}_{\omega} \boldsymbol{n},\\
    \boldsymbol{n} & = \mathop{argmax}\limits_{||\boldsymbol{n}||_2=1}  \alpha^2 \boldsymbol{n}^{T}\boldsymbol{J}_{\omega}^T \boldsymbol{J}_{\omega} \boldsymbol{n}. 
\end{aligned}
\end{equation}
As a result, the effective editing directions should be the eigenvectors of $\boldsymbol{J}_{\omega}^T \boldsymbol{J}_{\omega}$ associated with large eigenvalues. $\boldsymbol{J}_{\omega}^T \boldsymbol{J}_{\omega}$ can be formulated as $\boldsymbol{J}_{\omega}^T \boldsymbol{J}_{\omega}=\boldsymbol{L}^{*}+\boldsymbol{S}^{*}$, where $\boldsymbol{L}^{*}$ is the low-rank representation of $\boldsymbol{J}_{\omega}^T \boldsymbol{J}_{\omega}$ and $\boldsymbol{S}^{*} $ is the corrupted sparse matrix. Then we have matrix $\boldsymbol{V}$ consisting of attribute vectors through singular value decomposition (SVD) of $\boldsymbol{L}^*$ as follows:
\begin{equation}
    \boldsymbol{V}=\left[ \boldsymbol{v}_1,\boldsymbol{v}_2,...,\boldsymbol{v}_{d_\omega} \right].
\end{equation}
More details of low-rank factorization and Alternating Directions Method of Multipliers (ADMM) can be found in LowRankGAN \cite{zhu2021lowrankgan} and related works \cite{10.1561/2200000016,2013}. 

To obtain attribute vectors editing a single frame only, we first set the target frame to be edited as frame A and the other as frame B. Then we calculate the Jacobian matrix $\boldsymbol{J}_{\omega A},\boldsymbol{J}_{\omega B} \in \mathbb{R}^{d_x \times d_{\omega}}$ of each frame. $r_{a}$ and $r_{b}$ are used to represent the rank of the low-rank representation of $\boldsymbol{J}_{\omega A}^{T}\boldsymbol{J}_{\omega A},\boldsymbol{J}_{\omega B}^{T}\boldsymbol{J}_{\omega B}$. Correspondingly, $\boldsymbol{V}_{A}$ and $\boldsymbol{V}_{B}$ represent singular matrices containing attribute vectors of $\boldsymbol{J}_{\omega A}^{T}\boldsymbol{J}_{\omega A}$ and $\boldsymbol{J}_{\omega B}^{T}\boldsymbol{J}_{\omega B}$. Naturally, we have $r_a$ attribute vectors $\boldsymbol{v}_a$ in $\boldsymbol{V}_{A}$ that significantly influence frame A. However, they may have effects on frame B as well. Similarly, $\boldsymbol{V}_{B}$ can be divided into:
\begin{equation}
\begin{aligned}
    \boldsymbol{V}_{B_1}&=\left[\boldsymbol{v}_{1},\boldsymbol{v}_{2},...,\boldsymbol{v}_{r_b}\right], \\
    \boldsymbol{V}_{B_2}&=\left[\boldsymbol{v}_{r_b+1},\boldsymbol{v}_{r_b+2},...,\boldsymbol{v}_{d_\omega} \right].
\end{aligned}
\end{equation}
By projecting the attribute vectors $\boldsymbol{v}_{a}$ into the null space $\boldsymbol{V}_{B_2}$ of frame B associated with zero singular values as:

\begin{equation}
   \boldsymbol{p}_{A}=(\boldsymbol{I}-\boldsymbol{V}_{B_1}\boldsymbol{V}_{B_1}^{T})\boldsymbol{v}_{a}=\boldsymbol{V}_{B_2}\boldsymbol{V}_{B_2}^{T} \boldsymbol{v}_{a}, 
\end{equation}
where $\boldsymbol{I}$ represents the identity matrix, we get attribute vectors $\boldsymbol{p}_{A}$ that can edit frame A but barely influence frame B. Similarly, exchanging frame A and frame B, we have attribute vectors $\boldsymbol{p}_{B}$ editing frame B only. We define the attribute vectors editing the former frame $\tilde{\boldsymbol{x}}_{t}$ in the image pairs as backward motion codes $\boldsymbol{\omega}_{b}$ and the ones editing the latter frame ${\tilde{\boldsymbol{x}}_{t\pm k}}$ as forward motion codes $\boldsymbol{\omega}_{f}$ for clarity.

\subsection{Latent Code Generator}
\label{latentgenerator}
Having the pre-trained MotionStyleGAN and motion codes, we synthesize videos with an LSTM-based latent code generator generating latent code sequences under the guidance of conv3d-based video discriminators.

\textbf{Generator Architecture} 
As shown in Figure \ref{fig:temporal}, the proposed video generator MotionVideoGAN consists of a latent code generator $L_{\omega}$ producing latent code sequences based on an arbitrary content code $\boldsymbol{\omega}_{0}=G_M(\boldsymbol{z}_0)$ and motion codes $\boldsymbol{\omega}_{b}$, $\boldsymbol{\omega}_{f}$ for continuous video generation with a pre-trained MotionStyleGAN model. 

Specifically, the latent code generator $L_{\omega}$ generates coefficient matrices of motion codes utilizing LSTM-based networks \cite{6795963}. It consists of an LSTM encoder $LSTM_{enc}$ encoding the content code $\boldsymbol{\omega}_0$ and LSTM decoders $LSTM_{dec}$ outputting coefficient matrices $ \boldsymbol{M}_t$ of motion codes $\boldsymbol{\omega}_{b}$, $\boldsymbol{\omega}_{f}$ depending on hidden states $\boldsymbol{h}_{t-1}$ and cell states $\boldsymbol{c}_{t-1},t=1,2,...,n$ from former layers:

\begin{equation}
\begin{aligned}
      \boldsymbol{h}_0, \boldsymbol{c}_0 &= LSTM_{enc}(\boldsymbol{\omega}_0), \\
     \boldsymbol{M}_t,\boldsymbol{h}_t,\boldsymbol{c}_t &= LSTM_{dec}(\boldsymbol{h}_{t-1},\boldsymbol{c}_{t-1}), t=1,2,...,n.
\end{aligned}
\end{equation}

Then we can obtain latent code sequences for video generation as follows:
\begin{equation}
\begin{aligned}
    \boldsymbol{\omega}_0 &= G_M(\boldsymbol{z}_0), \\
    \boldsymbol{\omega}_t &= \boldsymbol{\omega}_{t-1} + \boldsymbol{M}_{t}\cdot\boldsymbol{\omega}_f,  (t=2k \leq n,k=1,2,...),  \\
     \boldsymbol{\omega}_t &= \boldsymbol{\omega}_{t-1} + \boldsymbol{M}_{t}\cdot\boldsymbol{\omega}_b, (t=2k+1 \leq n,k=0,1,...).\\
\end{aligned}    
\end{equation}
It is worth noting that the content code is manipulated with backward and forward motion codes in turn. We can generate image pairs as:
\begin{align}
    \left[\tilde{\boldsymbol{x}}_{t0},\tilde{\boldsymbol{x}}_{t1}\right]=G_S(\boldsymbol{\omega}_t).
\end{align}
Assuming that the latter image $\tilde{\boldsymbol{x}}_{01}$ generated from the content code $\boldsymbol{\omega}_0$ is chosen as the first frame, latent code $\boldsymbol{\omega}_1$ would be the linear combination of $\boldsymbol{\omega}_0$ and backward motion codes $\boldsymbol{\omega}_b$. In this way, the original former frame $\tilde{\boldsymbol{x}}_{00}$ is edited to $\tilde{\boldsymbol{x}}_{10}$ with motion codes $\boldsymbol{\omega}_b$ while the new latter frame $\tilde{\boldsymbol{x}}_{11}$ stays consistent with $\tilde{\boldsymbol{x}}_{01}$. Naturally, $\tilde{\boldsymbol{x}}_{10}$ becomes the second frame in the generated video. In the next step, latent code $\boldsymbol{\omega}_2$ would be the linear combination of $\boldsymbol{\omega}_1$ and forward motion codes $\boldsymbol{\omega}_f$. Correspondingly, $\tilde{\boldsymbol{x}}_{21}$ becomes the third frame. Repeating this process, we have the image sequence $\tilde{\boldsymbol{x}}_{01},\tilde{\boldsymbol{x}}_{10},\tilde{\boldsymbol{x}}_{21},\tilde{\boldsymbol{x}}_{30},...$ as the generated video. The video generation process is clearly exhibited in Figure \ref{fig:temporal} with images sharing green backgrounds and connected by orange arrows. We use mixed backgrounds for generated images $\tilde{\boldsymbol{x}}_{n0}$ and $\tilde{\boldsymbol{x}}_{n1}$ because $n$ can be an odd or even number. Since MotionStyleGAN is trained to learn bidirectional motions with image pairs, the choice of the first frame does not influence video generation effects. Suppose $\tilde{\boldsymbol{x}}_{00}$ is chosen as the first frame. In that case, we can exchange the order of forward and backward motion codes and get the image sequence $\tilde{\boldsymbol{x}}_{00},\tilde{\boldsymbol{x}}_{11},\tilde{\boldsymbol{x}}_{20},\tilde{\boldsymbol{x}}_{31},...$ as the generated video.

As illustrated in Section \ref{motioncode}, we only edit one of the images in image pairs generated every time step, ensuring that images in a video can share the same contents. When it comes to the video generation pipeline, we think it is more reasonable to edit the former and latter images with motion codes in turn. If we always keep one of the images unchanged and edit the other, motions generated in the following frames would be constrained since the MotionStyleGAN is trained with image pairs sampled from video datasets with fixed intervals. As a result, the generated videos may lack motion diversity and produce looped motions. 

\textbf{Video Discriminators}
    Prior works \cite{Tulyakov_2018_CVPR, tian2021a} utilize two discriminators operating on image and video levels to discriminate contents and motions in videos, respectively. However, our method only demands video-level discrimination due to the contents shared by generated image pairs of MotionStyleGAN. We apply conv3d-based video discriminators to the training of latent code generator $L_{\omega}$. We propose a bidirectional video discriminator $D_R$ in addition to the traditional video discriminator $D_V$. Traditional video discriminator $D_V$ learns from generated videos $\tilde{\boldsymbol{v}}$ labeled as fake and training videos $\boldsymbol{v}$ labeled as real. Letting $G_V$ denote the proposed video generator, $p_v$ and $p_z$ denote the distribution of real videos and normal distribution, the following adversarial loss is minimized in the training of $D_V$ and $L_{\omega}$:
\begin{equation}
\begin{aligned}
    \mathcal{L}_{D_V} &= \mathbb{E}_{\boldsymbol{v}\sim p_v}\left[log(D_V(\boldsymbol{v}))\right] \\
    &+ \mathbb{E}_{\boldsymbol{z}_0\sim p_{z}}\left[log(1-D_V(G_V(\boldsymbol{z}_0)))\right].
\end{aligned}
\end{equation}

It is elementary to discriminate between the reverse and original ones for regular videos. However, discrimination would become extremely difficult when it comes to looped videos. Therefore, we concatenate reverse videos with original videos as inputs for the bidirectional video discriminator $D_R$ in order to avoid generating looped videos. We define the proposed adversarial loss, named bidirectional adversarial loss, as follows:
\begin{equation}
\begin{aligned}
    \mathcal{L}_{D_R} &= \mathbb{E}_{\boldsymbol{v}\sim p_v}\left[log(D_R(\left[\boldsymbol{v},\boldsymbol{v}_r\right]))\right] \\
    &+ \mathbb{E}_{\boldsymbol{z}_0\sim p_{z}}\left[log(1-D_R(\left[G_V(\boldsymbol{z}_0),G_V(\boldsymbol{z}_0)_r\right]))\right],
\end{aligned}
\end{equation}
where $\boldsymbol{v}_r$ and $G_V(\boldsymbol{z}_0)_r$ represent reverse training and generated videos. The overall optimization target for latent code generator $L_{\omega}$ and video discriminators $D_V$ and $D_R$ is the combination of the two adversarial losses illustrated above: 
\begin{equation}
 \mathop{min}\limits_{L_{\omega}}(\mathop{max}\limits_{D_V}\mathcal{L}_{D_V}+\mathop{max}\limits_{D_R}\mathcal{L}_{D_R}).
\end{equation}

\section{Experiments}
As illustrated in Sec \ref{method}, MotionVideoGAN is a three-stage video generation method. We first evaluate the generation quality of the image pair generator MotionStyleGAN (Sec \ref{experiments}) and provide editing examples using motion codes (Sec \ref{motioncodeeval}). Based on the pre-trained MotionStyleGAN and calculated motion codes, we generate videos using the latent code generator as shown in Figure \ref{fig:temporal} and evaluate results compared with prior works (Sec \ref{videogeneration}). We employ three modern video datasets UCF101 $256^2$ \cite{soomro2012ucf101}, FaceForensics $256^2$ \cite{2018faceforensics}, and SkyTimelapse $256^2$ \cite{Xiong_2018_CVPR} for evaluation. Following the settings in StyleGAN-V \cite{stylegan_v}, we use train+test splits of the UCF101 dataset and train splits of the other two datasets for video generator training. All the datasets have 25 FPS.

\subsection{MotionStyleGAN}
\label{experiments}
The proposed MotionStyleGAN built on top of StyleGAN2 \cite{Karras_2020_CVPR} aims to generate image pairs sharing the same contents and producing various motions as $\left[ \tilde{\boldsymbol{x}}_{t}, \tilde{\boldsymbol{x}}_{t\pm k} \right] = G_P(\boldsymbol{z}_t)$. As demonstrated in Section \ref{method},  the training video datasets $\emph{D}=\left\lbrace \boldsymbol{x}_{t}\right\rbrace_{t=1}^{n}$ are reorganized as image pairs $\left[ \boldsymbol{x}_{t},\boldsymbol{x}_{t+k} \right]$ and $\left[ \boldsymbol{x}_{t+k},\boldsymbol{x}_{t} \right], 1\leq t,t+k\leq n$ with fixed frame interval $k$. For one thing, it would be hard for MotionStyleGAN to learn motions from extremely similar images with short intervals. For another, no effective motion information could be learned from image pairs with too long intervals. Therefore, we set the interval $k=4$ empirically for all three datasets in our experiments.

\begin{figure}[t]
\centering
\includegraphics[width=1.0\linewidth]{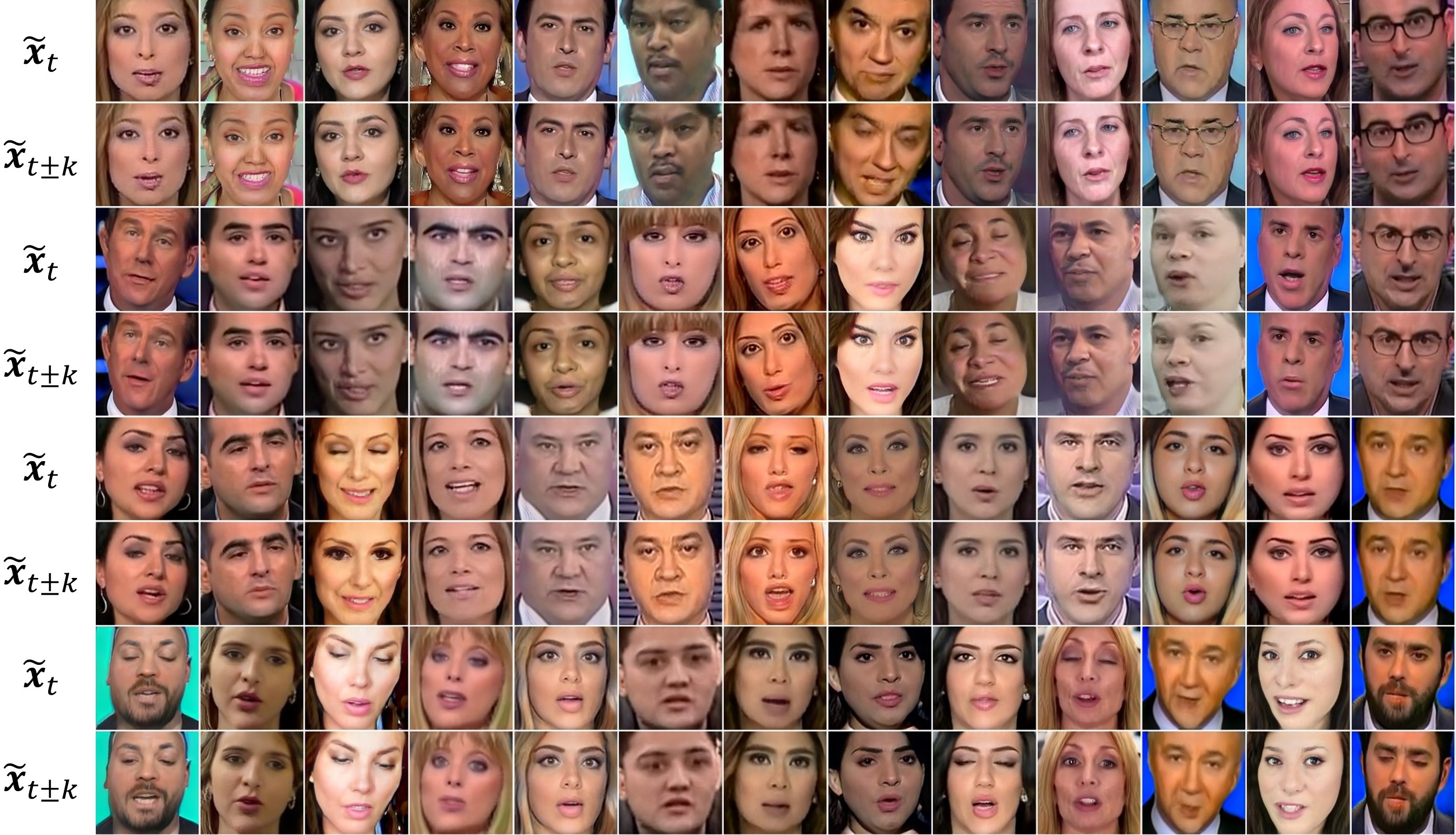}
\caption{Generated image pair examples by MotionStyleGAN trained on FaceForensics $256^2$. The former and latter frames are shown in two adjacent rows.}
\label{faceimage2}
\end{figure}

\begin{figure}[t]
\centering
\includegraphics[width=1.0\linewidth]{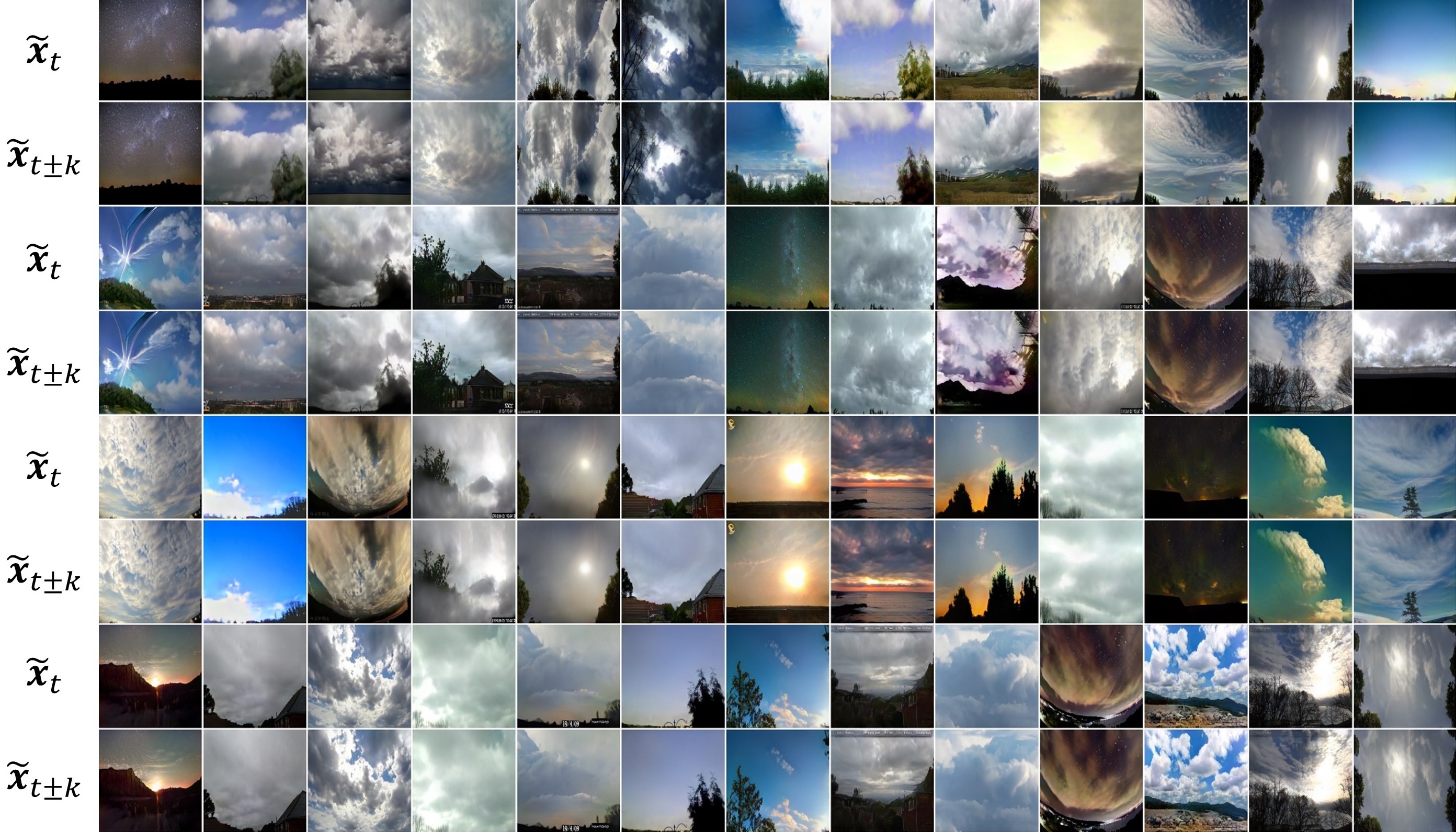}
\caption{Generated image pair examples by MotionStyleGAN trained on SkyTimelapse $256^2$. The former and latter frames are shown in two adjacent rows.}
\label{skyimage}
\end{figure}

FID \cite{10.5555/3295222.3295408} is employed to evaluate the image generation quality of MotionStyleGAN. We calculate FID with the former and latter frames separately and get almost the same results. We report FID values averaged over the former and latter frames in the following experiments. As shown in Table \ref{table:fid}, MotionStyleGAN achieves better FID results than StyleGAN2 used by MoCoGAN-HD \cite{tian2021a} on all three datasets. MotionStyleGAN generates high-quality image pairs that share the same contents and produce various motions by simultaneously learning contents and motions from image pairs sampled from video datasets. In Figure \ref{faceimage2} and \ref{skyimage}, we present generation examples of the MotionStyleGAN models trained on FaceForensics $256^2$  and SkyTimelapse $256^2$. Typical motion patterns can be found in generated image pairs, including winking and talking from FaceForensics $256^2$ and cloud moving from SkyTimelapse $256^2$.

\begin{table}[t]
\caption{FID ($\downarrow$) of the proposed MotionStyleGAN on UCF101 $256^2$, FaceForensics $256^2$, and SkyTimelapse $256^2$. The training was done on $\times 4$ 24 GB NVidia GeForce RTX 3090 GPUs. Results are compared with StyleGAN2 used by MoCoGAN-HD \cite{tian2021a}.}
\centering
\begin{tabular}{l|c|c|c}
\hline
Approach & UCF101  & FaceForensics & SkyTimelapse \\
\hline
StyleGAN2 & 45.63 & 10.99 & 10.80  \\
MotionStyleGAN & $\boldsymbol{37.70}$ & $\boldsymbol{9.65}$ & $\boldsymbol{10.48}$\\
\hline
\end{tabular}
\label{table:fid}
\end{table}

MotionStyleGAN shares approximate parameters scale and training speed with StyleGAN2. Taking image resolution $256 \times 256$ as an example, MotionStyleGAN has only 9237 more parameters than StyleGAN2, while both models have roughly $58.9M$ parameters. We provide more detailed comparison of computational costs in Appendix (Sec \ref{appendixb}). We train MotionStyleGAN on $\times 4$ 24 GB NVidia GeForce RTX 3090 GPUs with $25000k$ image pairs. Then the checkpoint with the lowest FID value is chosen for further motion codes calculation and latent code generator training. 

\subsection{Motion Code}
\label{motioncodeeval}
\begin{figure}[t]
\centering
\includegraphics[width=1.0\linewidth]{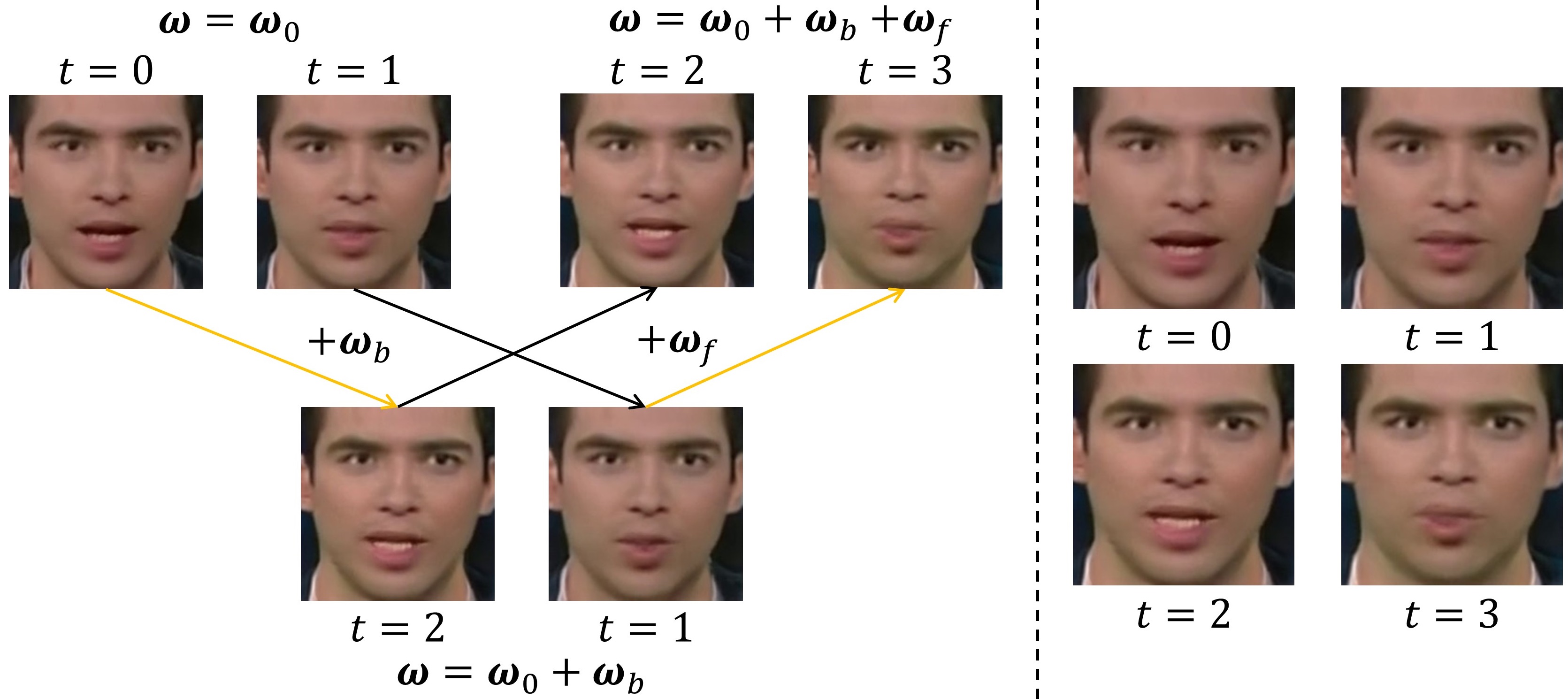}
\caption{Image editing examples using motion codes on FaceForensics $256^2$. \textbf{Left}: The image pair on the top left is generated by MotionStyleGAN from the content code $\boldsymbol{\omega}_0$. The former frame is edited with backward motion codes $\boldsymbol{\omega}_b$ as shown in the image pair below. Then the latter frame is edited with forward motion codes $\boldsymbol{\omega}_f$ as shown on the top right. Yellow arrows represent editing with motion codes while black arrows represent no apparent changes.  \textbf{Right}: A short image sequence generated by manipulating the content code with motion codes. Here $t$ only represents the order, not a specific time or time step.}
\label{fig:motion1}
\end{figure}

As shown in Figure \ref{fig:motion1}, image pairs generated by MotionStyleGAN can be edited with motion codes obtained through the low-rank factorization of MotionStyleGAN's Jacobian matrix. The former and latter frames can be edited with backward and forward motion codes $\boldsymbol{\omega}_b, \boldsymbol{\omega}_f$ without affecting the other, respectively. We can maintain the contents of the edited frame by keeping the other frame stable when editing its motions.

\subsection{Video Generation}
\label{videogeneration}

\textbf{Evaluation Metrics}
Frechet Video Distance (FVD) has no official implementation and is greatly influenced by how many clips are chosen per video in datasets despite its wide application to evaluate distribution distance between two video datasets. We follow the project implemented by StyleGAN-V \cite{stylegan_v} to evaluate FVD by picking a random clip per real video to compare with 2048 generated videos. Considering that videos in datasets generally have uneven video length distributions (see Table \ref{table:dataset} in Appendix (Sec \ref{appendixb})), different training strategies would influence the FVD metric. MoCoGAN-HD and the proposed approach are based on pre-trained image (pair) generators. Therefore, we are more likely to generate videos similar to the longer ones since the image (pair) generators are pre-trained on all frames from video datasets, which would hurt the FVD results under the above protocol. We do not view this as a shortcoming since long videos provide more information than shorter ones. If we pick as many clips as possible per video, it would be unfair for approaches trained on videos directly (a random clip per video per epoch). Therefore, we split approaches based on pre-trained image (pair) generators (MoCoGAN-HD and ours) and approaches trained directly on videos (MoCoGAN, StyleGAN-V, et al.) into two groups for fair FVD comparison as shown in Table \ref{table:UCF101} and \ref{table:fvd}. 

\begin{table}[htbp]
\caption{FVD ($\downarrow$) and IS ($\uparrow$) of  MotionVideoGAN compared with other video generation approaches on UCF101 $256^2$. The training was done on $\times 4$ 24 GB NVidia GeForce RTX 3090 GPUs. Video generation approaches are split into two groups based on different training strategies.}
\centering
\resizebox{1.0\columnwidth}{!}{
\begin{tabular}{l|c|c}
\hline
Approach & FVD ($\downarrow$) & IS ($\uparrow$)\\
\hline
MoCoGAN (CVPR 2018)  & 2886.9 & $10.09 \pm 0.30$ \\
VideoGPT (Arxiv) & 2880.6 & $12.61 \pm 0.33$ \\
DIGAN (ICLR 2022) & 1630.2 & $23.16 \pm 1.13$ \\
StyleGAN-V (CVPR 2022) & $\boldsymbol{1431.0}$ & $23.94 \pm 0.73$ \\
\cline{1-2}
MoCoGAN-HD (ICLR 2021) & 1729.6 & $23.39 \pm 1.48$ \\
MotionVideoGAN (ours)  & $\boldsymbol{1031.4}$ & $\boldsymbol{25.88 \pm 0.39}$\\
\hline
\end{tabular}}
\label{table:UCF101}
\end{table}

\begin{table}[htbp]
\caption{FVD ($\downarrow$) of MotionVideoGAN compared with other video generation approaches on FaceForensics $256^2$ and SkyTimelapse $256^2$. Video generation approaches are split into two groups based on different training strategies.}
\centering
\begin{tabular}{l|c|c}
\hline
Approach & FaceForensics  & SkyTimelapse \\
\hline
MoCoGAN (CVPR 2018) & 124.7 & 206.6 \\
VideoGPT (Arxiv) & 185.9 & 222.7 \\
DIGAN (ICLR 2022) & 62.5 & 83.1 \\
StyleGAN-V (CVPR 2022) & $\boldsymbol{47.4}$ & $\boldsymbol{79.5}$ \\
\hline
MoCoGAN-HD (ICLR 2021) & 111.8 & 164.1\\
MotionVideoGAN (ours)  & $\boldsymbol{108.8}$ & $\boldsymbol{157.9}$ \\
\hline
\end{tabular}
\label{table:fvd}
\end{table}

\begin{figure}[t]
\centering
\includegraphics[width=1.0\linewidth]{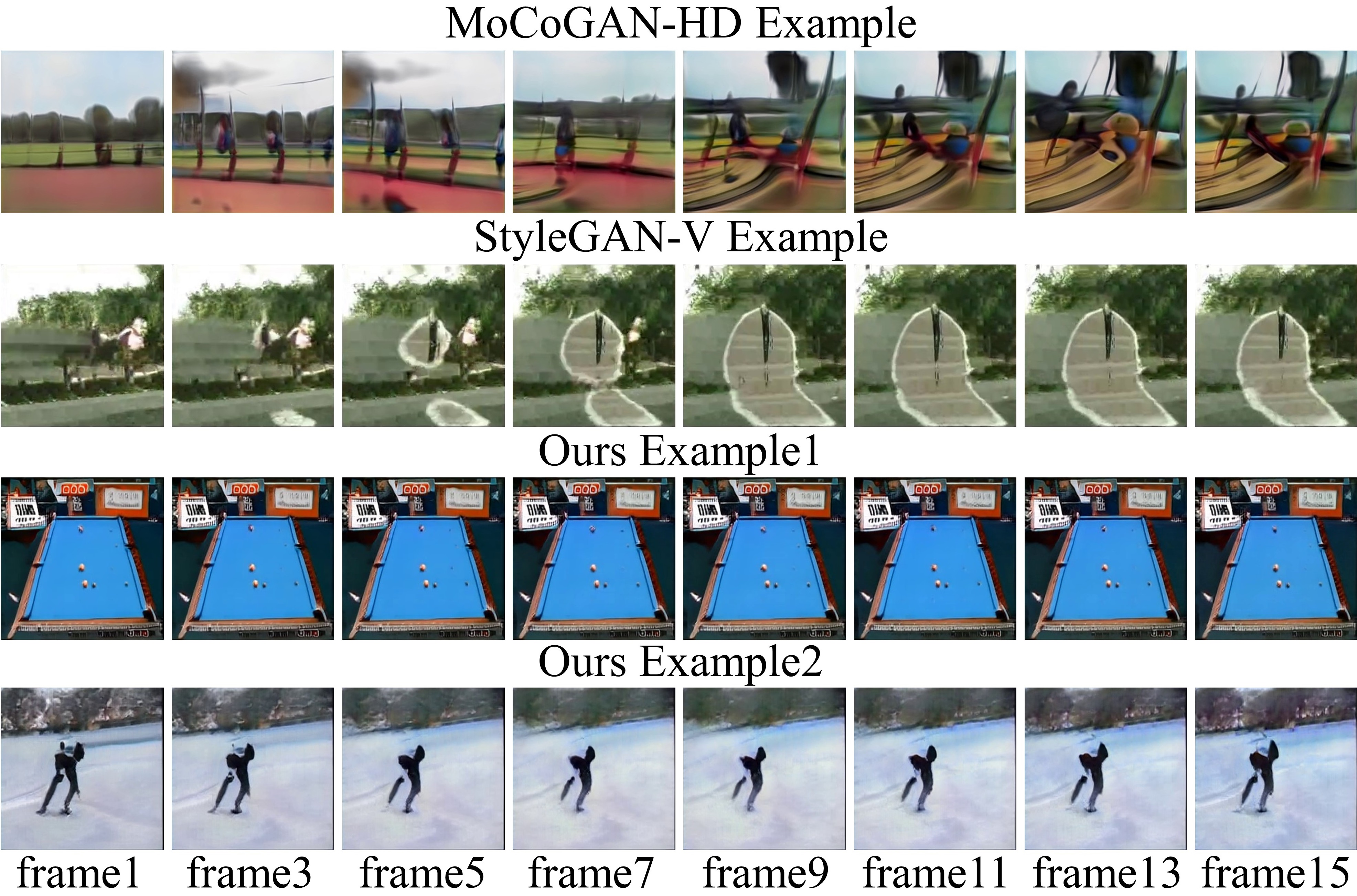}
\caption{Generated video examples of 16 frames by MotionVideoGAN trained on UCF101 $256^2$. We can synthesize videos of various scenes with consistent contents. For comparison, videos synthesized by MoCoGAN-HD and StyleGAN-V are shown in the top two rows.}
\label{fig:ucfvideo}
\end{figure}

We employ Inception Score (IS) to evaluate video generators on UCF101 $256^2$ using a C3D model fine-tuned on UCF101 \cite{TGAN}. Unlike FVD, IS is not influenced by different training strategies. 

\textbf{Video Generation Evaluation} Based on the pre-trained MotionStyleGAN model and calculated motion codes, we train the latent code generator to generate videos of 16 frames following settings in prior works. We first present results on the complex dataset UCF101, which contains 13320 videos from 101 categories. Our approach achieves state-of-the-art results of FVD and IS on UCF101 $256^2$ as shown in Table \ref{table:UCF101}. Despite the influence of uneven video length distribution, the proposed MotionVideoGAN achieves a more than $27\%$ reduction of FVD value benefiting from the content consistency in complex scenes achieved with the motion space learned by MotionStyleGAN. We compare the visualization results of our approach with MoCoGAN-HD and StyleGAN-V on UCF101 $256^2$ in Figure \ref{fig:ucfvideo}. Our approach achieves more plausible videos than previous works benefiting from the consistent contents.

\begin{figure}[htbp]
\centering
\includegraphics[width=1.0\linewidth]{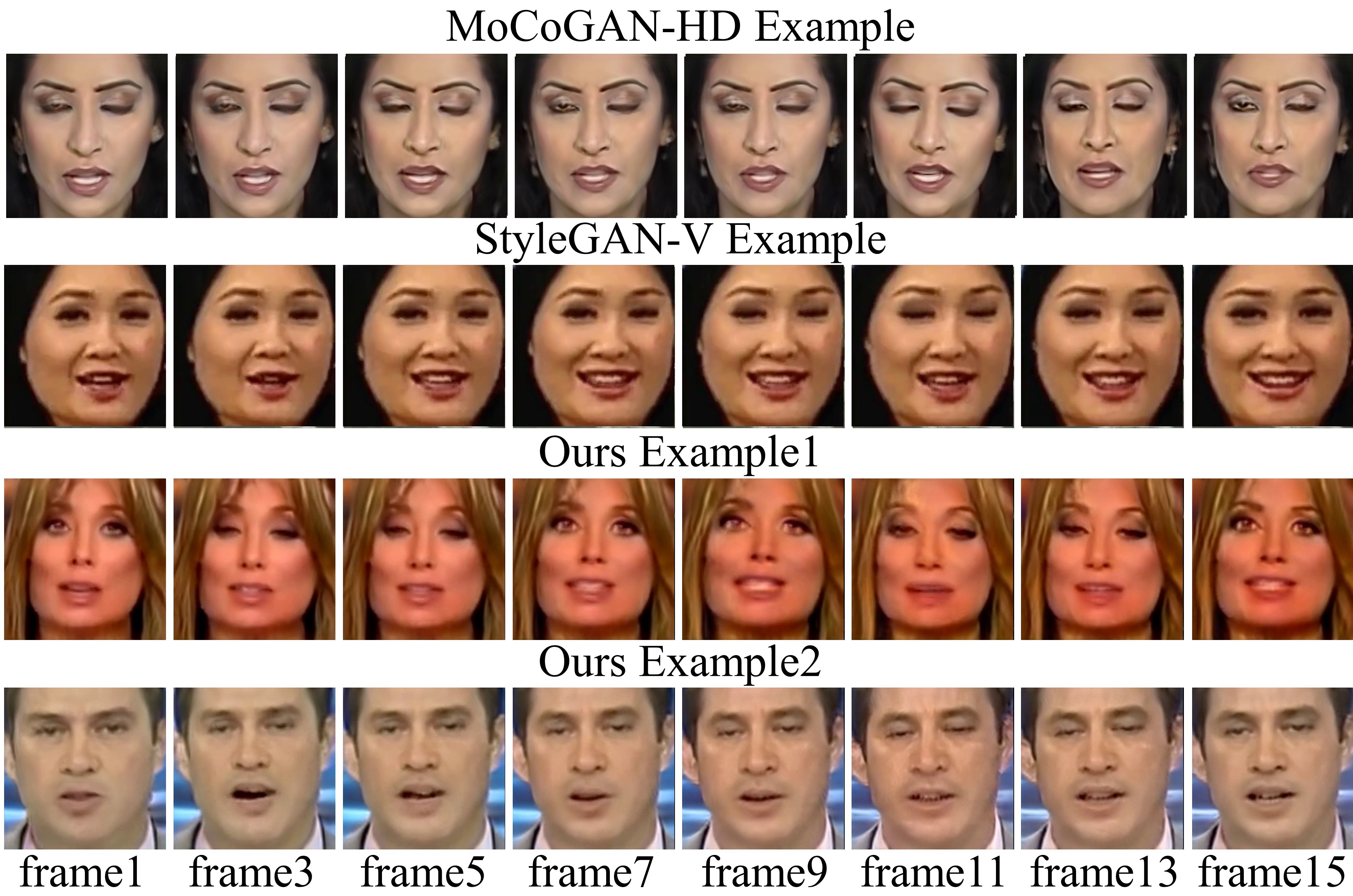}
\caption{Generated video examples of 16 frames by MotionVideoGAN trained on FaceForensics $256^2$.  For comparison, videos synthesized by MoCoGAN-HD and StyleGAN-V are shown in the top two rows.}
\label{fig:facevideo}
\end{figure}

\begin{figure}[htbp]
\centering
\includegraphics[width=1.0\linewidth]{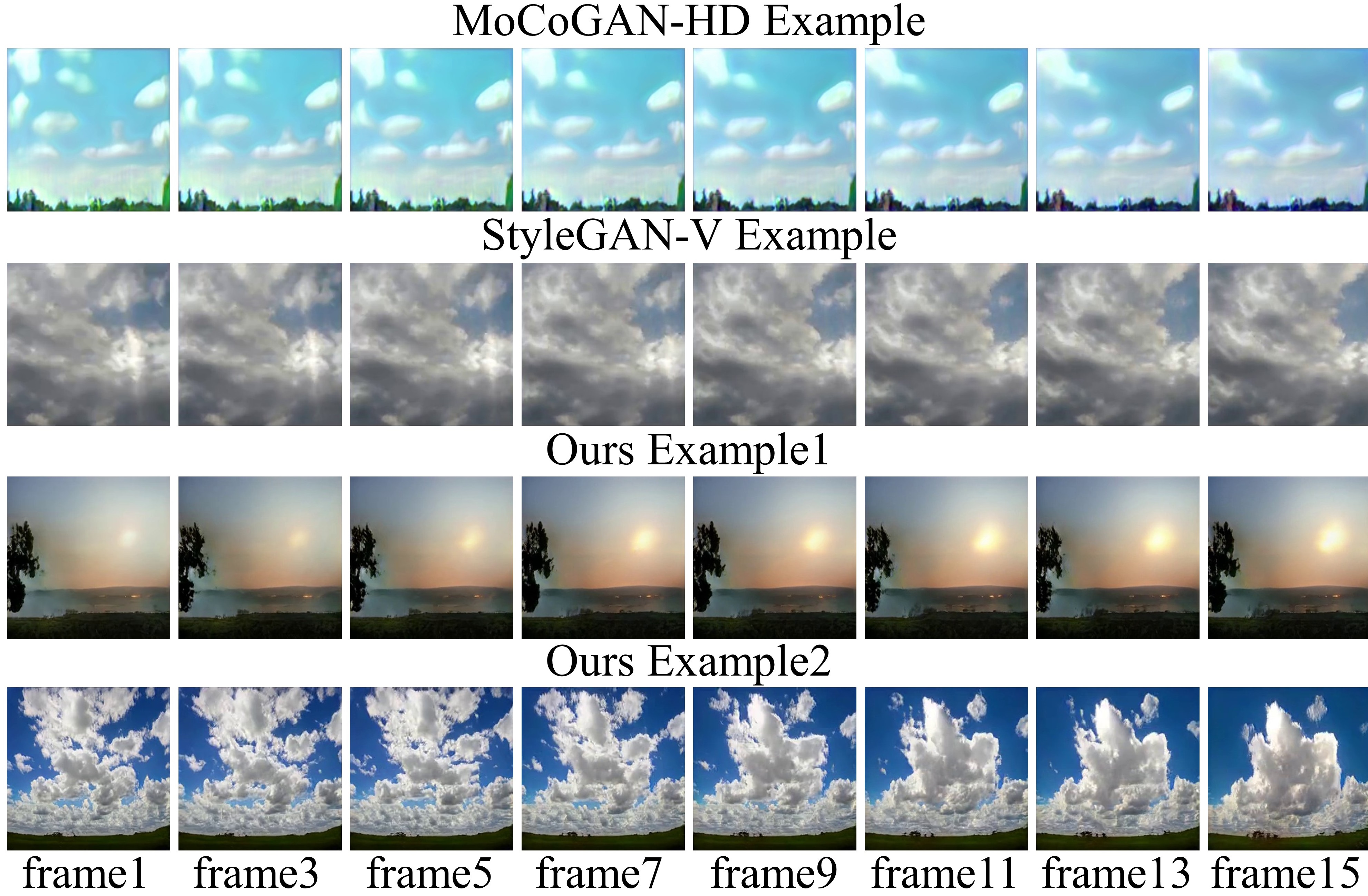}
\caption{Generated video examples of 16 frames by MotionVideoGAN trained on SkyTimelapse $256^2$. For comparison, videos synthesized by MoCoGAN-HD and StyleGAN-V are shown in the top two rows.}
\label{fig:skyvideo}
\end{figure}

As for FaceForensics $256^2$ and SkyTimelapse $256^2$ containing 704 and 2376 videos of a single category, our approach achieves better FVD than MoCoGAN-HD, which shares the same strategy of using pre-trained image (pair) generators with us. Influenced by the uneven video length distributions (see Table \ref{table:dataset} in Appendix (Sec \ref{appendixb})), we obtain relatively higher FVD values than StyleGAN-V and DIGAN, which are trained directly on videos. However, our approach can still generate realistic videos as visualized in Figure $\ref{fig:facevideo}$ and $\ref{fig:skyvideo}$. More video generation examples of our approach can also be found in Appendix (Sec \ref{appendixc}). 

\begin{figure*}[t]
\centering
\includegraphics[width=1.0\linewidth]{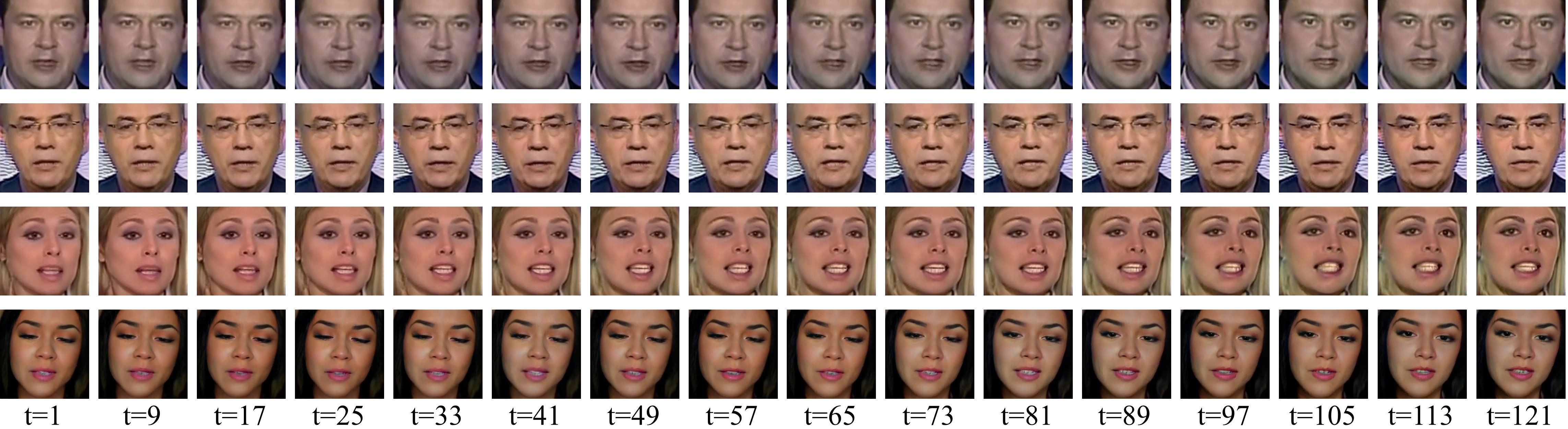}
\caption{Generated video examples of 128 frames by MotionVideoGAN trained with 16-frame clips on FaceForensics $256^2$. We sample several 128-frame video examples and display each 8-th frame, starting from the first frame (t=1). We provide long videos generated based on the latent code generator trained with original videos and subsampled videos in the top two rows and the bottom two rows, respectively.}
\label{fig:longvideo}
\end{figure*}

\textbf{Long Sequence Generation}
The proposed MotionVideoGAN video generator is constrained by limited computational resources and trained to generate videos of 16 frames following prior works. We can synthesize long videos through interpolation in the latent space. Based on the latent code generator trained with original videos, longer videos changing more slowly can be obtained with interpolated latent code sequences. As a supplement, we train the latent code generator with subsampled videos and interpolate the latent code sequences to synthesize long videos with a similar temporal evolution speed to training datasets. The proposed video generator achieves stable results of long sequence generation when trained with subsampled videos, benefiting from the MotionStyleGAN trained with image pairs having intervals. Figure \ref{fig:longvideo} visualizes the synthesized long videos of 128 frames on FaceForensics $256^2$ by MotionVideoGAN. Long videos generated based on the latent code generator trained with original videos and subsampled videos are shown in the top two rows and the bottom two rows, respectively.

\begin{table}[t]
\caption{Ablations of discriminators (loss functions) on FaceForensics $256^2$. ID, TVD, BVD represents the image discriminator, traditional and bidirectional video discriminator, respectively. $\mathcal{L_{D_I}}$ represents the adversarial loss of the image discriminator.}
\centering
\begin{tabular}{c|c|c}
\hline
Discriminator Settings & Loss & FVD ($\downarrow$) \\
\hline
TVD & $\mathcal{L_{D_V}}$ & 143.7\\
BVD & $\mathcal{L_{D_R}}$ &149.4 \\
BVD+TVD+ID & $\mathcal{L_{D_V}}+\mathcal{L_{D_R}}+\mathcal{L_{D_I}}$  & 112.6\\
BVD+TVD (ours) & $\mathcal{L_{D_V}}+\mathcal{L_{D_R}}$ & $\boldsymbol{108.8}$\\
\hline
\end{tabular}
\label{table:ablation}
\end{table}

\begin{table}[t]
\caption{Ablation analysis of frame interval $k$ on FaceForensics $256^2$.}
\begin{center}
\begin{tabular}{c|c|c|c|c}
\hline
Frame Interval $k$ & 2 & 3 & 4 & 5\\
\hline
FVD ($\downarrow$) & 190.5 & 134.6 & $\boldsymbol{108.8}$ & 160.3\\
\hline
\end{tabular}
\end{center}
\label{table:ablationk}
\end{table}

\textbf{Fast Convergence}
Our approach can naturally maintain content consistency benefiting from the learned motion space. Therefore, we achieve faster convergence than sequential generation methods like MoCoGAN-HD, which uses pre-trained image generators and randomly sampled motion codes. The proposed MotionVideoGAN only needs 45 epochs to achieve better FVD than MoCoGAN-HD trained for 800 epochs on FaceForensics $256^2$ without increasing the scale of parameters massively. As for UCF101 $256^2$, we train the MotionVideoGAN for 5 epochs and obtain superior promotion of FVD than MoCoGAN-HD trained for 20 epochs. Considering the complexity of video datasets, our approach can significantly save training costs.

\subsection{Ablations}
\textbf{Ablation Analysis of Discriminators}
We train MotionVideoGAN on FaceForensics $256^2$ under different settings of discriminators corresponding to different loss functions. The FVD results are reported in Table \ref{table:ablation}. Except for the video discriminators illustrated in Sec \ref{latentgenerator}, we also employ the image discriminator \cite{tian2021a}, which uses the first frame of generated videos as real samples and the following frames as fake samples for ablation analysis. We train models for 45 epochs under different settings of discriminators for fair comparison. The full MotionVideoGAN approach using two video discriminators outperforms other settings, indicating the importance of each video discriminator. We find that the addition of the image discriminator provides no better results.  Since our approach generates videos within the motion space and achieves content consistency, the image discriminator is no longer needed and may influence the motion learning and content consistency of the proposed approach, leading to slower convergence.

\begin{figure}[t]
\centering
\includegraphics[width=1.0\linewidth]{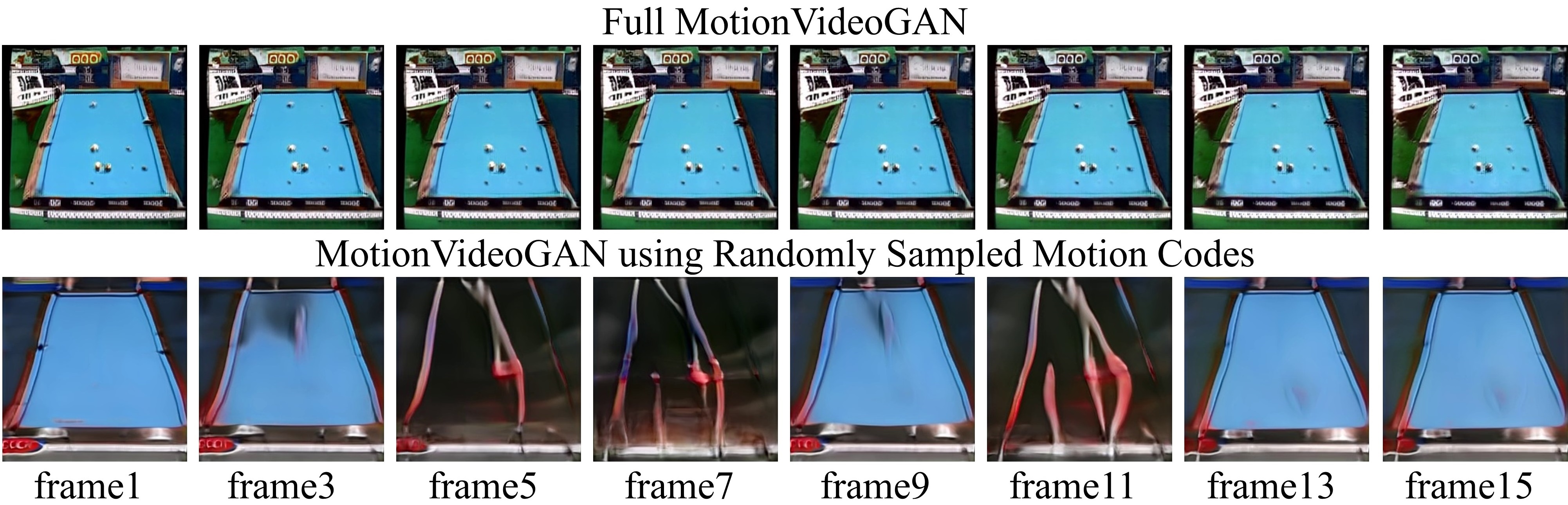}
\caption{Video generation examples on UCF101 $256^2$ produced by full MotionVideoGAN and MotionVideoGAN using randomly sampled motion codes.}
\label{fig:ablationcode}
\end{figure}

\textbf{Ablation Analysis of Frame Interval}
\label{appendixk}
The proposed MotionStyleGAN is trained on image pairs with fixed frame intervals to learn motions and build the motion space to realize content consistency for video generation. We carry out ablations of frame interval on FaceForensics $256^2$ \cite{2018faceforensics} as shown in Table \ref{table:ablationk}. For too small intervals, it is hard for MotionStyleGAN to learn motion changes. As a result, the proposed video generator has a higher probability of generating videos close to static. As for too-long intervals, effective motions would be difficult to learn, leading to discontinuous results. Considering that all the employed datasets share the same FPS of 25, we apply the frame interval of 4 to all of them in our experiments.

\textbf{Ablation Analysis of Motion Codes}
We add experiments of our approach using randomly sampled codes instead of the calculated motion codes for video generation on the complex dataset UCF101. Using randomly sampled motion codes leads to seriously worse FVD results: 1578.3, compared with the SOTA result of 1031.4 achieved by our approach. We add visualized comparison in Fig. \ref{fig:ablationcode}. It can be seen that content consistency is hard to maintain with randomly sampled motion codes.

\section{Limitations}
\textbf{Limitations of Video Continuity} The video generator MotionVideoGAN is based on the pre-trained image pair generator MotionStyleGAN. Although the MotionVideoGAN can generate high-quality and continuous videos in most cases, we find it difficult for the MotionVideoGAN to produce continuous results under certain circumstances. Since the MotionStyleGAN is trained with image pairs sampled from video datasets with fixed intervals, it may be lacking in learning the motions in small ranges, leading to discontinuous generated videos. 

\begin{figure}[t]
\centering
\includegraphics[width=1.0\linewidth]{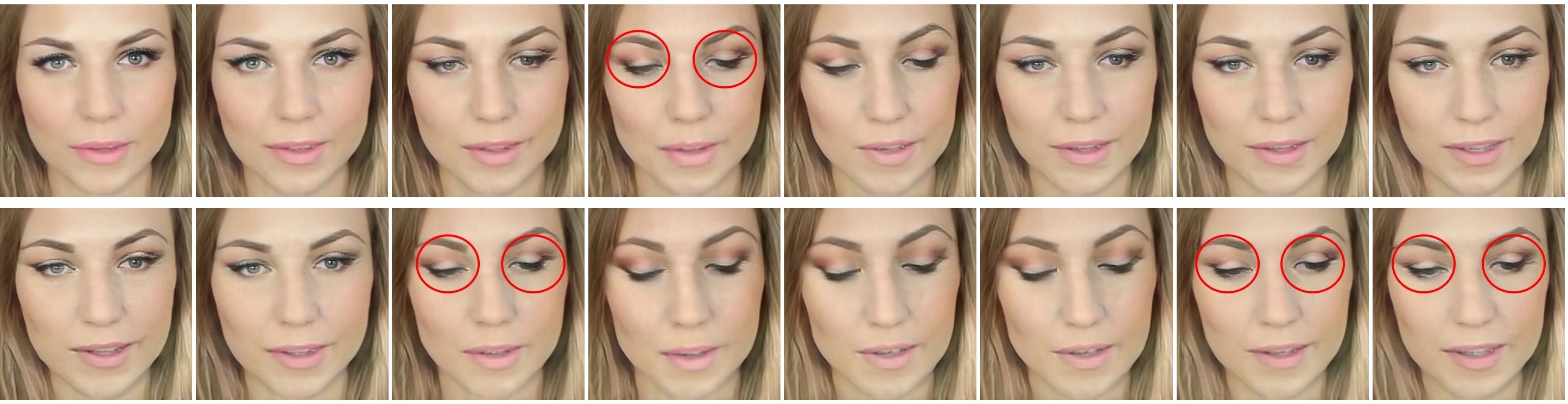}
\caption{A video generation example (16 consecutive frames) on FaceForensics $256^2$ produced by MotionVideoGAN. (top row: frame 1-8, bottom row: frame 9-16)}
\label{fig:motion}
\end{figure}

We try to further improve continuity by adding image pairs with shorter intervals ($k=2,3$) and provide an example in Figure \ref{fig:motion} (optimized FVD $97.4$ on FaceForensics $256^2$ \cite{2018faceforensics}), where more continuous details like half-opened eyes (circled out) can be found. We will work on more effective methods to further improve the continuity of our approach in future research.

\textbf{Limitations of Video Length} The MotionVideoGAN produces image sequences by editing the content code with motion codes. Similar to prior works, it is hard for our generator to produce hours-long videos since too many steps of editing may bias the edited latent codes away from the latent space, resulting in poor generation quality. Therefore, future research is needed to keep the edited latent codes in the latent space for hours-long video generation.

\section{Conclusion}
This work presents a novel video generator MotionVideoGAN to synthesize videos based on the motion space learned from image pairs. More specifically, we first propose an image pair generator MotionStyleGAN to generate high-quality image pairs sharing the same contents and producing various motions. Then we develop the MotionVideoGAN using an LSTM-based latent code generator to obtain latent code sequences and generate videos based on the pre-trained MotionStyleGAN model and motion codes obtained through low-rank factorization of the MotionStyleGAN's Jacobian matrix. 

Our work provides a novel method to build the motion space for image editing and video generation with image pair generators. To the best knowledge of us, the proposed MotionVideoGAN is the first to synthesize videos based on motions learned by pre-trained image pair generators. Moreover, our approach achieves state-of-the-art performance and superior promotion on the complex UCF101 $256^2$ dataset. In this paper, we implement our approach based on StyleGAN2. However, it is worth noting that our approach is not constrained to a specific model and can be applied to more powerful image-based GANs in the future for better results.  

\section{Appendix}

\subsection{More Details of Proposed Models}
\label{appendixa}

\textbf{MotionStyleGAN}
We propose the image pair generator MotionStyleGAN built on top of StyleGAN2 \cite{Karras_2020_CVPR} to generate image pairs sharing the same contents and producing various motions. We modify the official Pytorch implementation of StyleGAN2 from three output color channels (RGB) for a single image to six color channels for image pairs. Although StyleGAN3 \cite{Karras2021} has been proposed to generate images that are fully equivariant to translation and rotation, we still choose StyleGAN2 as the baseline model of MotionStyleGAN since we find it difficult for StyleGAN3 to achieve equivariance of translation and rotation on the employed video datasets. For future work, we plan to use StyleGAN3 for large-resolution human face video datasets like MEAD $1024^2$ \cite{kaisiyuan2020mead}. 

We train the proposed MotionStyleGAN on UCF101 $256^2$ \cite{soomro2012ucf101}, FaceForensics $256^2$ \cite{2018faceforensics}, and SkyTimelapse $256^2$ \cite{Xiong_2018_CVPR} with gamma (R1 regularization weight, used in StyleGAN models) value of 1 and ADA augmentation method \cite{NEURIPS2020_8d30aa96}.  The training was done on $\times 4$ 24 GB NVidia GeForce RTX 3090 GPUs. The learning rates of the generator and discriminator are set as 0.002, equally. The batch size for each dataset is set as 32, 16, and 32. In our experiments, we find it more suitable to use a smaller batch size for small-scale datasets like FaceForensics $256^2$, which contains 704 training videos. While for large-scale datasets like UCF101 $256^2$ and SkyTimelapse $256^2$, a larger batch size is needed for faster convergence.

\textbf{Motion Code}
We reimplement the program for motion codes with Pytorch based on the official TensorFlow implementation provided by LowRankGAN \cite{zhu2021lowrankgan}. The former and latter images in image pairs are set as the frames to be edited to generate of backward and forward motion codes, respectively.

\textbf{MotionVideoGAN}
In the proposed MotionVideoGAN, both the LSTM encoder and decoder have input and hidden sizes of 512, the same as the latent codes in MotionStyleGAN. Trainable parameter matrices are applied to transform the output of LSTM decoders to coefficient matrices $M_t$ of motion codes, as shown in Figure \ref{fig:temporal}.

As for the conv3d-based video discriminators, we follow the network architecture proposed in MoCoGAN-HD \cite{tian2021a} which has input sequences of six color channels. The traditional video discriminator takes videos concatenated with repeated initial images as inputs. The proposed bidirectional video discriminator shares the same network architecture but concatenates the original and reverse videos as inputs. 

The proposed MotionVideoGAN is trained to generate videos on UCF101 $256^2$, FaceForensics $256^2$, and SkyTimelapse $256^2$ based on pre-trained MotionStyleGAN models and corresponding motion codes. We select 30 motion codes for video generation and set the learning rate of the latent code generator as 0.0001. The MotionVideoGAN is trained for 5 epochs, 45 epochs, and 20 epochs on UCF101 $256^2$, FaceForensics $256^2$, and SkyTimelapse $256^2$, respectively. 

\subsection{More Details of Model Evaluation}
\label{appendixb}

\textbf{MotionStyleGAN Evaluation}
As reported in Table \ref{table:fidappendix}, we achieve close FID results with less training kimgs (kimg: thousand image (pair)) compared with StyleGAN2 \cite{Karras_2020_CVPR} used in MoCoGAN-HD \cite{tian2021a} on FaceForensics $256^2$ \cite{2018faceforensics} and SkyTimelapse $256^2$ \cite{Xiong_2018_CVPR}. Additionally, MoCoGAN-HD employs SkyTimelapse $128^2$ for evaluation. Our approach achieves a lower FID value on larger resolution. As for UCF101 $256^2$  \cite{soomro2012ucf101}, the MotionStyleGAN achieves an apparently lower FID value with close training kimgs. We report the FID results to prove that MotionStyleGAN can generate high-quality images benefiting from the learned motion space and provide evaluation details as a reference.

\begin{table*}
\caption{FID results and corresponding training kimgs of the proposed MotionStyleGAN on UCF101 $256^2$, FaceForensics $256^2$, and SkyTimelapse $256^2$. Results are compared with StyleGAN2 used by MoCoGAN-HD \cite{tian2021a}.}
\begin{center}
\renewcommand{\arraystretch}{1.05}
\begin{tabular}{l|cccccc}
\hline
Method & \multicolumn{2}{c}{UCF101} & \multicolumn{2}{c}{FaceForensics}  & \multicolumn{2}{c}{SkyTimelapse} \\ \cline{2-7}
& FID ($\downarrow$) & kimg &  FID ($\downarrow$) & kimg &  FID ($\downarrow$) & kimg \\ 
\hline
StyleGAN2 & 45.63 & $\boldsymbol{6935}$ & 10.99 & 8765 & 10.80 & 12633  \\
MotionStyleGAN  & $\boldsymbol{37.70}$ & 7680 & $\boldsymbol{9.65}$ & $\boldsymbol{4800}$ & $\boldsymbol{10.48}$ & $\boldsymbol{8600}$\\
\hline
\end{tabular}
\end{center}
\label{table:fidappendix}
\end{table*}

\begin{table*}
\caption{Additional statistical information of datasets, including the number of videos, max video length, mean video length, and variance of video length. Video length is measured in terms of the number of frames.}
\begin{center}
\begin{tabular}{l|c|c|c|c}
\hline
Datasets & Number of Videos & Max Length & Mean Length & Variance \\
\hline
UCF101  & 13320 & 1776 & 186.50 & 9556.995 \\
FaceForensics  & 704 & 1957 & 517.07 & 46867.212 \\
SkyTimelapse  & 2376 & 266309 & 493.01 & 30794675.650 \\
\hline
\end{tabular}
\end{center}
\label{table:dataset}
\end{table*}

\begin{figure}[t]
\centering
\includegraphics[width=1.0\linewidth]{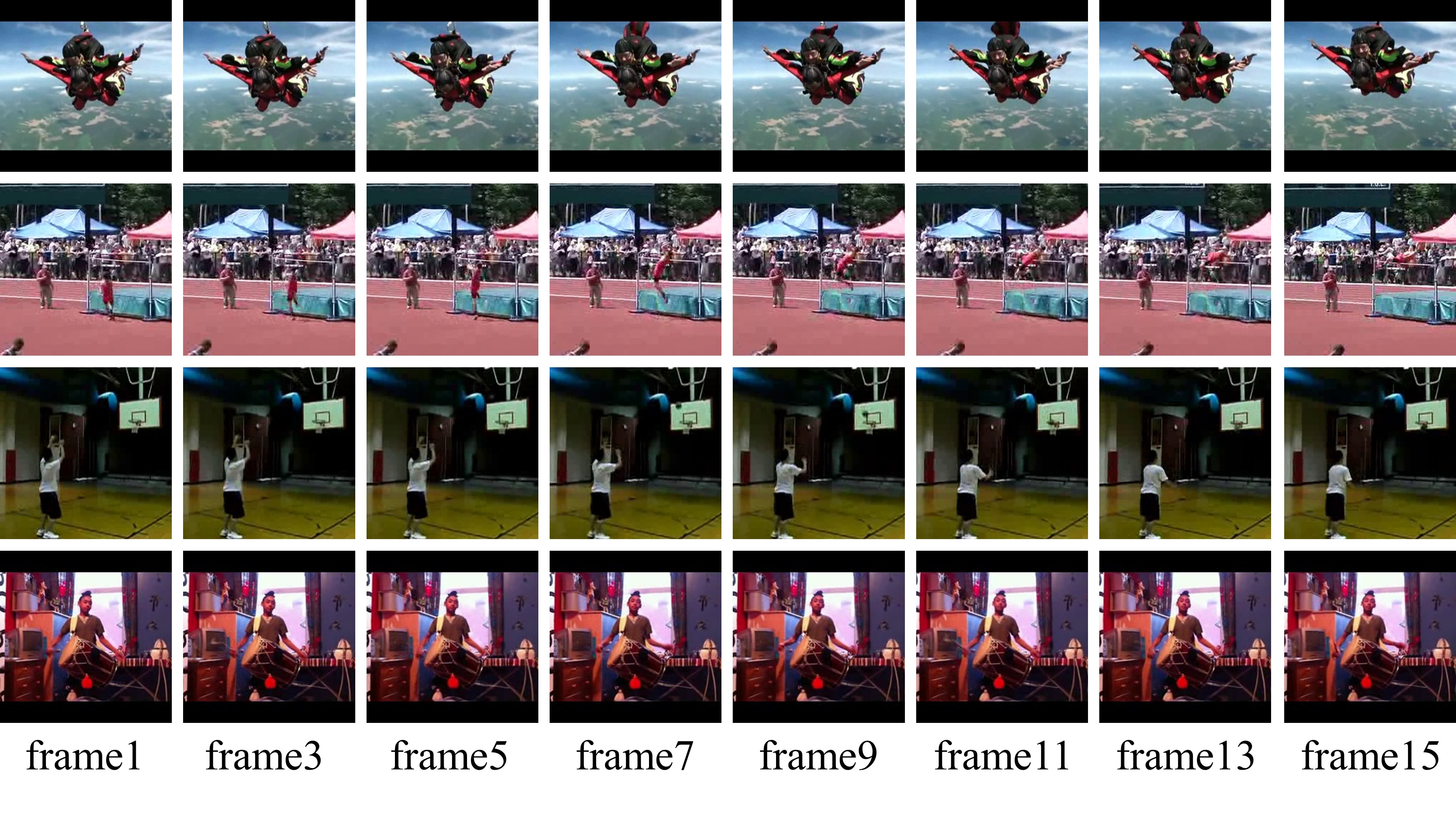}
\caption{Example videos of 16 frames randomly picked from UCF101 $256^2$.}
\label{fig:ucfvideoexample}
\end{figure}

\begin{figure}[t]
\centering
\includegraphics[width=1.0\linewidth]{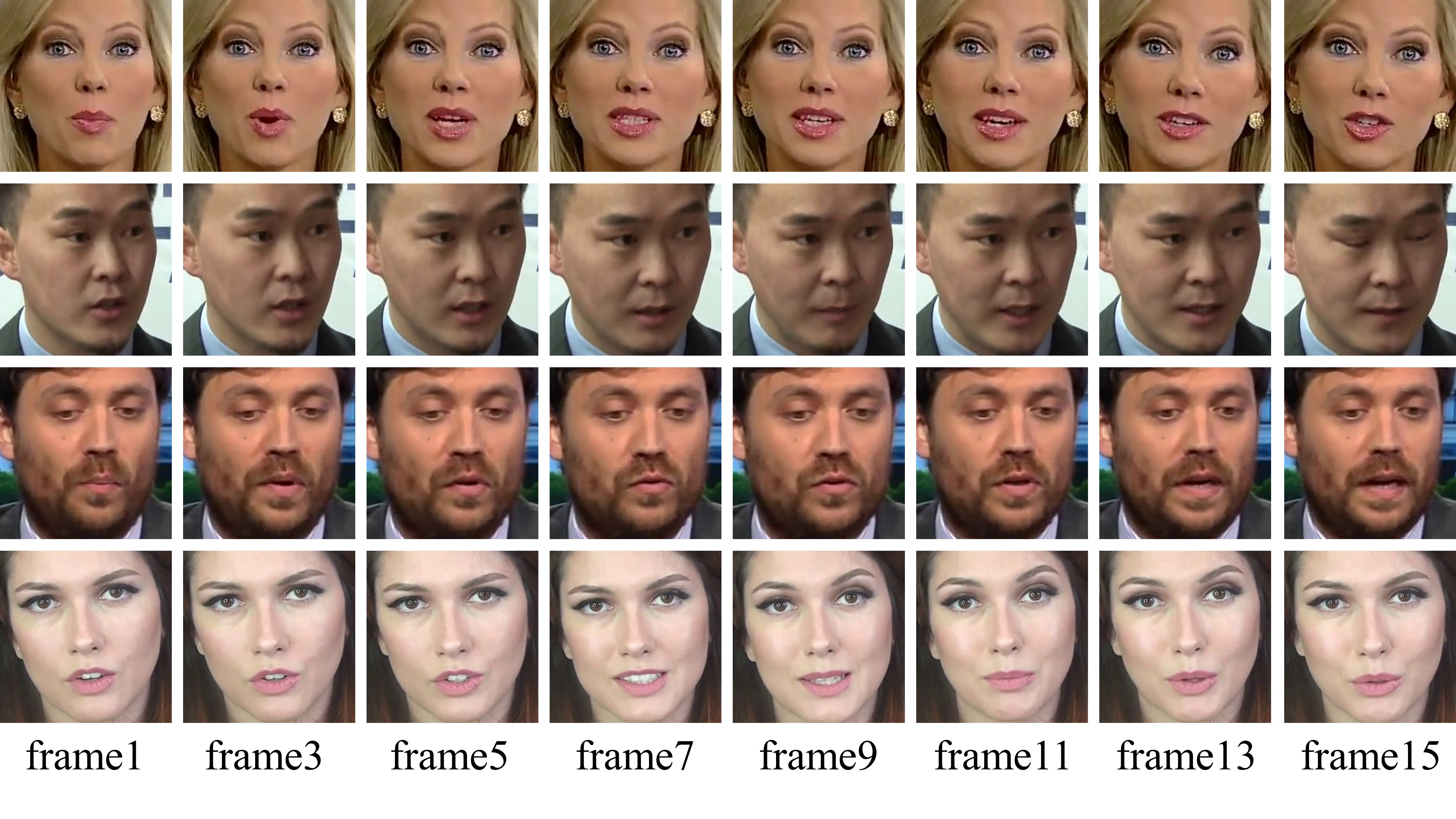}
\caption{Example videos of 16 frames randomly picked from FaceForensics $256^2$.}
\label{fig:facevideoexample}
\end{figure}

\begin{figure}[t]
\centering
\includegraphics[width=1.0\linewidth]{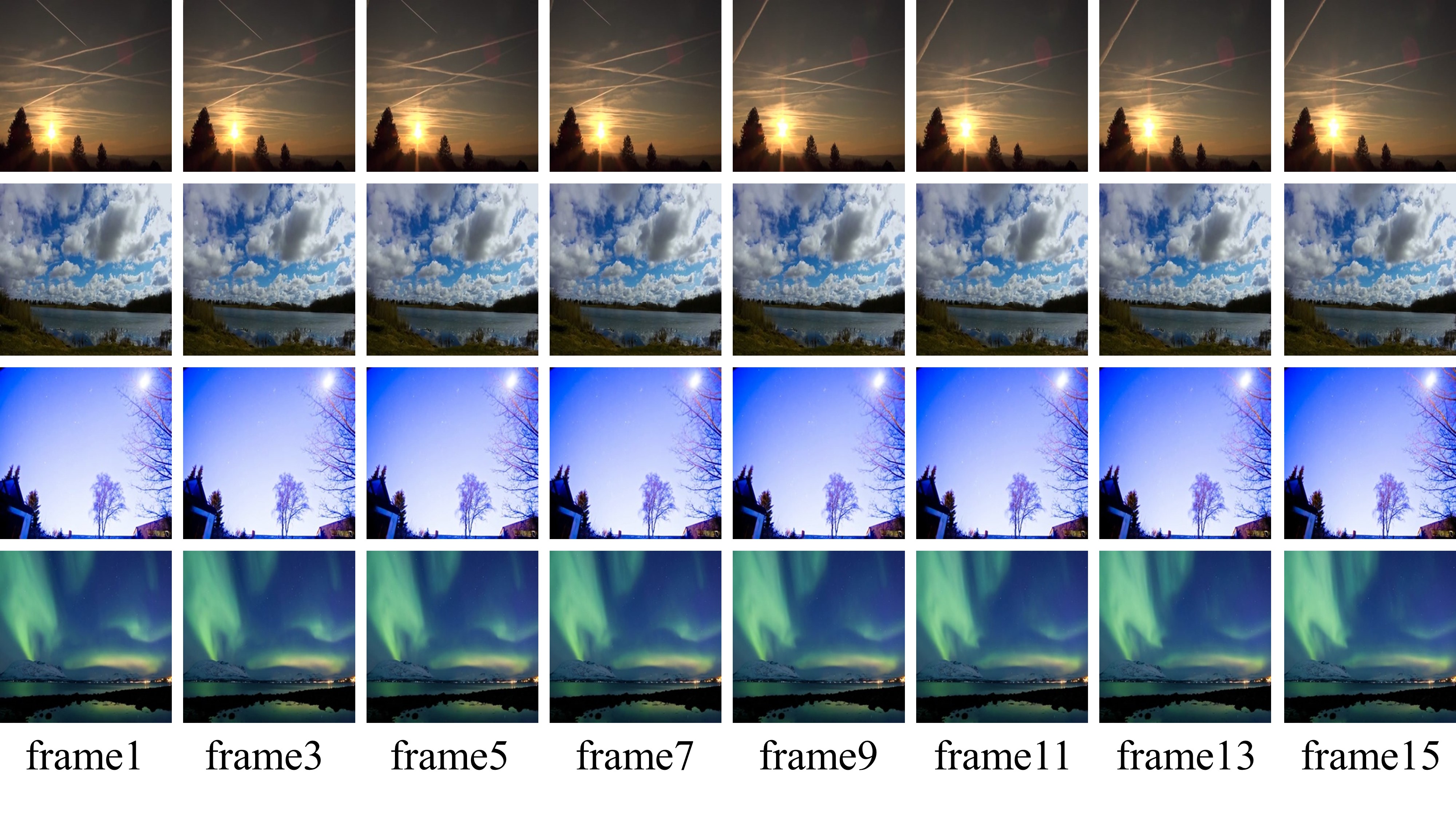}
\caption{Example videos of 16 frames randomly picked from SkyTimelapse $256^2$.}
\label{fig:skyvideoexample}
\end{figure}

\begin{figure}[t]
\centering
\includegraphics[width=1.0\linewidth]{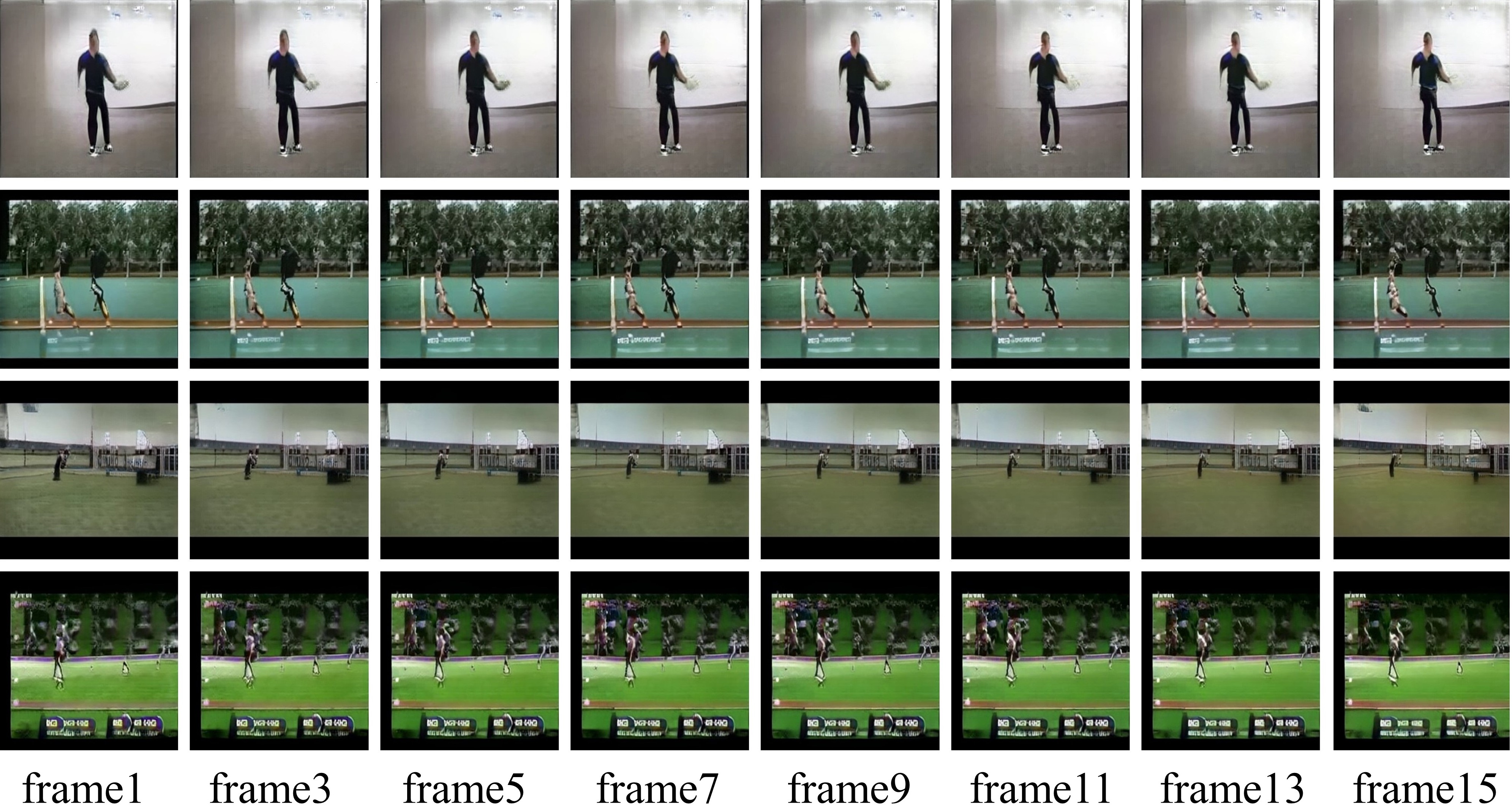}
\caption{Supplementary generated video examples of 16 frames by MotionVideoGAN trained on UCF101 $256^2$.}
\label{fig:ucfvideoadd}
\end{figure}

\textbf{MotionVideoGAN Evaluation}
 We provide the statistics of used datasets in Table \ref{table:dataset}, including the number of videos, max video length, mean video length, and variance of video length. Based on the variance of video length, it can be seen that the video length distribution of the UCF101 dataset is more concentrated than the other two datasets. 
 
 The uneven distributions of video length hurt the FVD results of our approach, especially on FaceForensics $256^2$ and SkyTimelapse $256^2$. However, our approach is still capable of generating high-quality videos on these two datasets (see Appendix (Sec \ref{appendixc})) although we cannot achieve state-of-the-art FVD results under the protocol of FVD stated in Sec \ref{videogeneration} influenced by the uneven video length distributions.

 MoCoGAN-HD based on pre-trained image generators suffers similar problems of FVD evaluation with our approach. To compare our approach with MoCoGAN-HD fairly, we directly use the original datasets without clipping long videos. Considering that our approach has already achieved high-quality results on these two datasets, it is not necessary to adjust the training strategy of the proposed MotionVideoGAN or clip video datasets only to pursue better results of the FVD metric.

\subsection{More Examples}
\label{appendixc}
In this section, we provide some typical examples from the used datasets and more generation results of our approach.

\begin{figure}[t]
\centering
\includegraphics[width=1.0\linewidth]{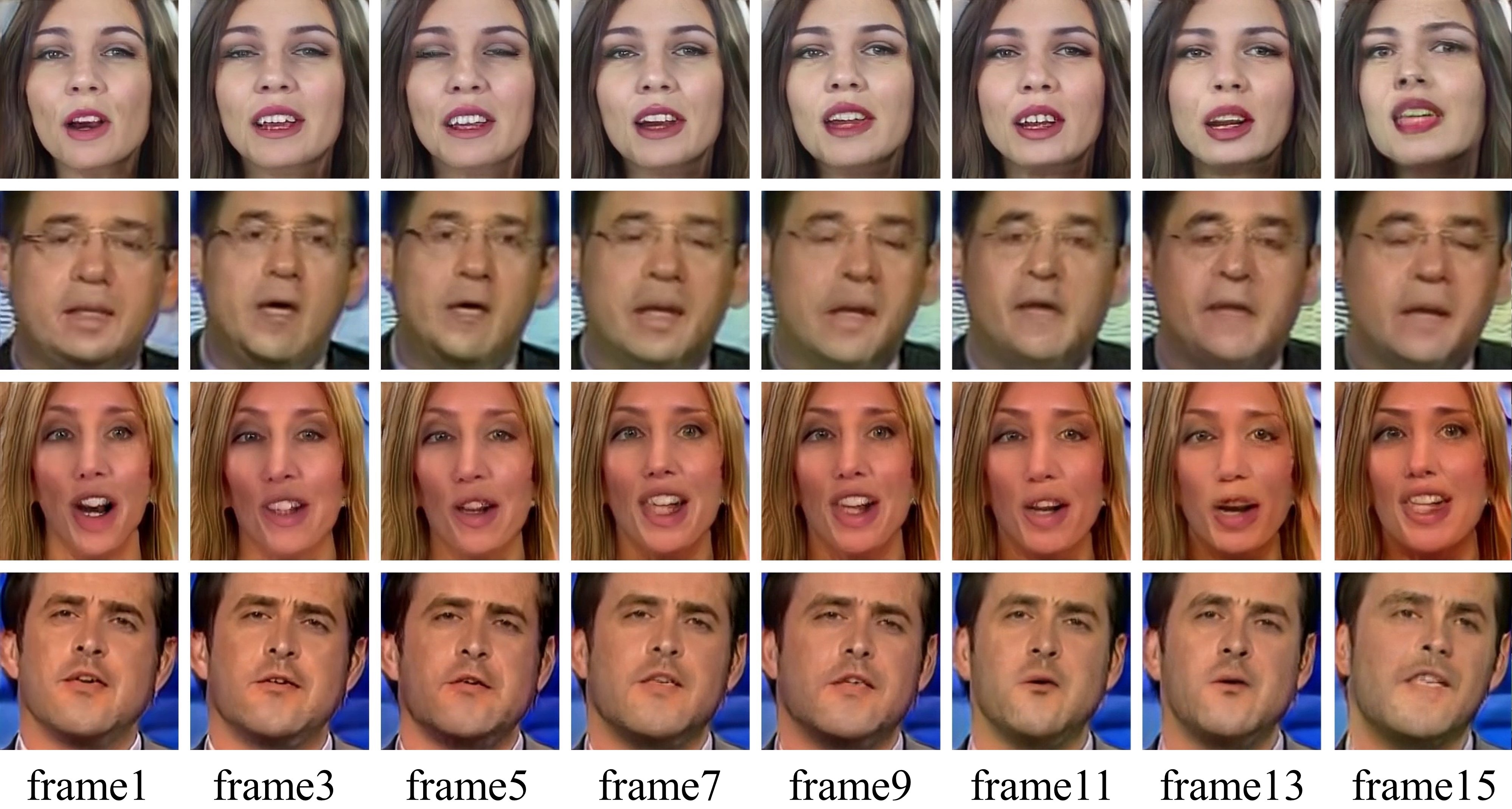}
\caption{Supplementary generated video examples of 16 frames by MotionVideoGAN trained on FaceForensics $256^2$.}
\label{fig:facevideoadd}
\end{figure}

\begin{figure}[t]
\centering
\includegraphics[width=1.0\linewidth]{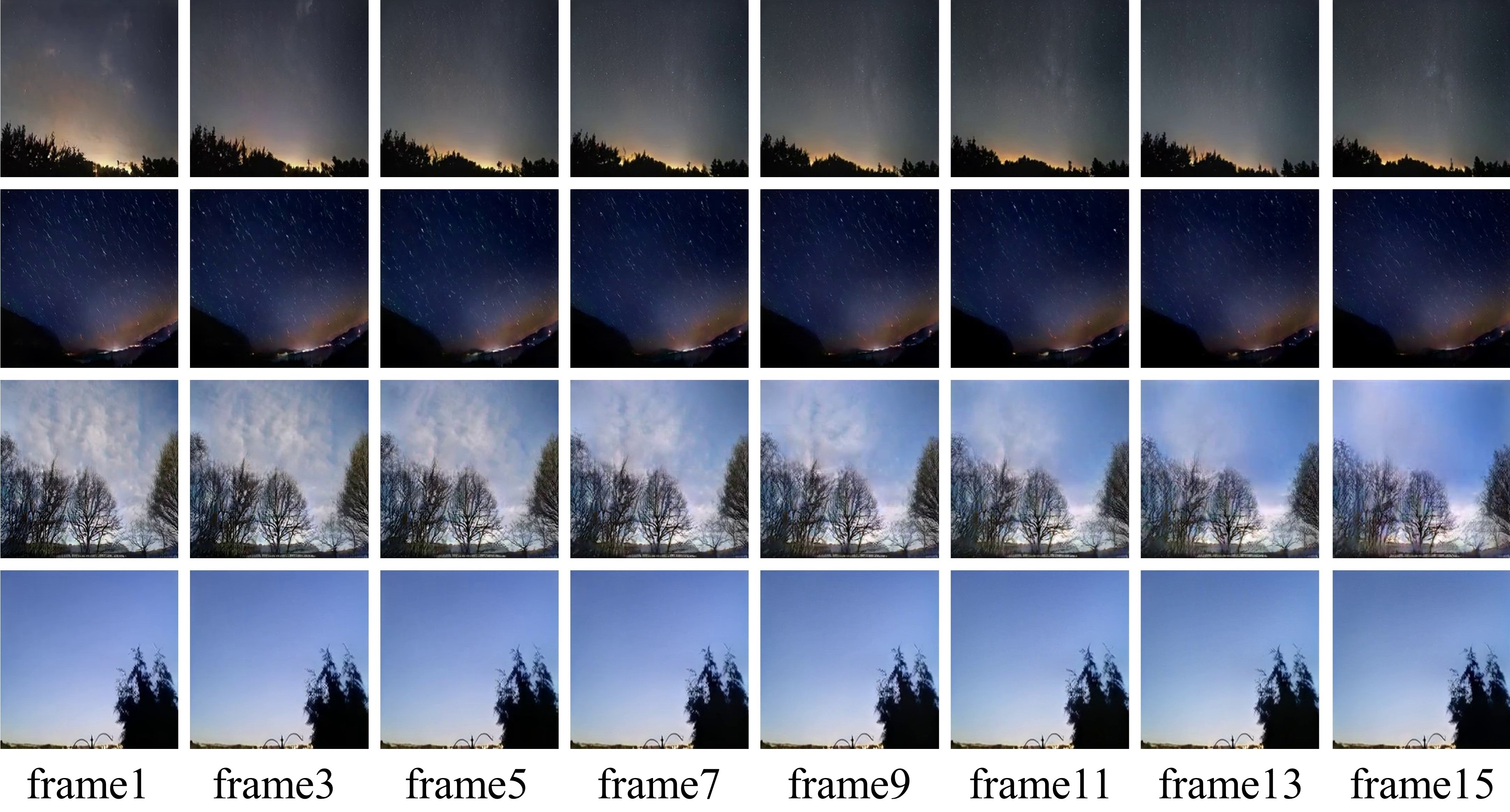}
\caption{Supplementary generated video examples of 16 frames by MotionVideoGAN trained on SkyTimelapse $256^2$.}
\label{fig:skyvideoadd}
\end{figure}

\textbf{Training Video Examples} Figure \ref{fig:ucfvideoexample}, \ref{fig:facevideoexample}, and \ref{fig:skyvideoexample} show some videos randomly picked from training video datasets used in this paper, UCF101 $256^2$ \cite{soomro2012ucf101}, FaceForensics $256^2$ \cite{2018faceforensics}, and SkyTimelapse $256^2$ \cite{Xiong_2018_CVPR}.

\textbf{Additional Examples of MotionVideoGAN}
More generation examples of the proposed video generator MotionVideoGAN on UCF101 $256^2$, FaceForensics $256^2$, and SkyTimelapse $256^2$ are shown in Figure \ref{fig:ucfvideoadd}, \ref{fig:facevideoadd}, and \ref{fig:skyvideoadd}.

\bibliographystyle{IEEEtran}
\bibliography{tmm}

\begin{thebibliography}{10}
\providecommand{\url}[1]{#1}
\csname url@samestyle\endcsname
\providecommand{\newblock}{\relax}
\providecommand{\bibinfo}[2]{#2}
\providecommand{\BIBentrySTDinterwordspacing}{\spaceskip=0pt\relax}
\providecommand{\BIBentryALTinterwordstretchfactor}{4}
\providecommand{\BIBentryALTinterwordspacing}{\spaceskip=\fontdimen2\font plus
\BIBentryALTinterwordstretchfactor\fontdimen3\font minus
  \fontdimen4\font\relax}
\providecommand{\BIBforeignlanguage}[2]{{%
\expandafter\ifx\csname l@#1\endcsname\relax
\typeout{** WARNING: IEEEtran.bst: No hyphenation pattern has been}%
\typeout{** loaded for the language `#1'. Using the pattern for}%
\typeout{** the default language instead.}%
\else
\language=\csname l@#1\endcsname
\fi
#2}}
\providecommand{\BIBdecl}{\relax}
\BIBdecl

\bibitem{Karras_2019_CVPR}
T.~Karras, S.~Laine, and T.~Aila, ``A style-based generator architecture for
  generative adversarial networks,'' in \emph{Proceedings of the IEEE/CVF
  Conference on Computer Vision and Pattern Recognition}, 2019, pp. 4401--4410.

\bibitem{Karras_2020_CVPR}
T.~Karras, S.~Laine, M.~Aittala, J.~Hellsten, J.~Lehtinen, and T.~Aila,
  ``Analyzing and improving the image quality of stylegan,'' in
  \emph{Proceedings of the IEEE/CVF Conference on Computer Vision and Pattern
  Recognition}, 2020, pp. 8110--8119.

\bibitem{Karras2021}
T.~Karras, M.~Aittala, S.~Laine, E.~H{\"a}rk{\"o}nen, J.~Hellsten, J.~Lehtinen,
  and T.~Aila, ``Alias-free generative adversarial networks,'' \emph{Advances
  in Neural Information Processing Systems}, vol.~34, pp. 852--863, 2021.

\bibitem{karras2018progressive}
T.~Karras, T.~Aila, S.~Laine, and J.~Lehtinen, ``Progressive growing of {GAN}s
  for improved quality, stability, and variation,'' in \emph{International
  Conference on Learning Representations}, 2018.

\bibitem{pmlr-v97-zhang19d}
H.~Zhang, I.~Goodfellow, D.~Metaxas, and A.~Odena, ``Self-attention generative
  adversarial networks,'' in \emph{International Conference on Machine
  Learning}.\hskip 1em plus 0.5em minus 0.4em\relax PMLR, 2019, pp. 7354--7363.

\bibitem{9721642}
Z.~Zheng, Y.~Bin, X.~Lu, Y.~Wu, Y.~Yang, and H.~T. Shen, ``Asynchronous
  generative adversarial network for asymmetric unpaired image-to-image
  translation,'' \emph{IEEE Transactions on Multimedia}, 2022.

\bibitem{pang2021image}
Y.~Pang, J.~Lin, T.~Qin, and Z.~Chen, ``Image-to-image translation: Methods and
  applications,'' \emph{IEEE Transactions on Multimedia}, 2021.

\bibitem{isola2017image}
P.~Isola, J.-Y. Zhu, T.~Zhou, and A.~A. Efros, ``Image-to-image translation
  with conditional adversarial networks,'' in \emph{Proceedings of the IEEE
  Conference on Computer Vision and Pattern Recognition}, 2017, pp. 1125--1134.

\bibitem{lee2018diverse}
H.-Y. Lee, H.-Y. Tseng, J.-B. Huang, M.~Singh, and M.-H. Yang, ``Diverse
  image-to-image translation via disentangled representations,'' in
  \emph{Proceedings of the European Conference on Computer Vision}, 2018, pp.
  35--51.

\bibitem{zhu2017unpaired}
J.-Y. Zhu, T.~Park, P.~Isola, and A.~A. Efros, ``Unpaired image-to-image
  translation using cycle-consistent adversarial networks,'' in
  \emph{Proceedings of the IEEE International Conference on Computer Vision},
  2017, pp. 2223--2232.

\bibitem{ma2018exemplar}
L.~Ma, X.~Jia, S.~Georgoulis, T.~Tuytelaars, and L.~Van~Gool, ``Exemplar guided
  unsupervised image-to-image translation with semantic consistency,''
  \emph{arXiv preprint arXiv:1805.11145}, 2018.

\bibitem{9583266}
X.~Tian, J.~Shao, D.~Ouyang, and H.~T. Shen, ``Uav-satellite view synthesis for
  cross-view geo-localization,'' \emph{IEEE Transactions on Circuits and
  Systems for Video Technology}, vol.~32, no.~7, pp. 4804--4815, 2022.

\bibitem{9807935}
J.~Vaishnavi and V.~Narmatha, ``Video captioning based on image captioning as
  subsidiary content,'' in \emph{2022 Second International Conference on
  Advances in Electrical, Computing, Communication and Sustainable Technologies
  (ICAECT)}, 2022, pp. 1--6.

\bibitem{8805998}
S.~K. Adari, W.~Garcia, and K.~Butler, ``Adversarial video captioning,'' in
  \emph{2019 49th Annual IEEE/IFIP International Conference on Dependable
  Systems and Networks Workshops (DSN-W)}, 2019, pp. 24--27.

\bibitem{111}
J.~Gu, L.~Liu, P.~Wang, and C.~Theobalt, ``Stylenerf: A style-based 3d aware
  generator for high-resolution image synthesis,'' in \emph{In International
  Conference on Learning Representations}, 2022.

\bibitem{222}
K.~Schwarz, Y.~Liao, M.~Niemeyer, and A.~Geiger, ``Graf: Generative radiance
  fields for 3d-aware image synthesis,'' in \emph{In Advances in Neural
  Information Processing Systems}, 2020.

\bibitem{333}
E.~R. Chan, C.~Z. Lin, M.~A. Chan, K.~Nagano, B.~Pan, S.~D. Mello, O.~Gallo,
  L.~Guibas, J.~Tremblay, S.~Khamis, T.~Karras, and G.~Wetzstein, ``Efficient
  geometry-aware 3d generative adversarial networks,'' in \emph{In Proceedings
  of the IEEE/CVF Conference on Computer Vision and Pattern Recognition}, 2022.

\bibitem{444}
E.~R. Chan, M.~Monteiro, P.~Kellnhofer, J.~Wu, and G.~Wetzstein, ``Pi-gan:
  Periodic implicit generative adversarial networks for 3d-aware image
  synthesis.'' in \emph{In Proceedings of the IEEE/CVF Conference on Computer
  Vision and Pattern Recognition}, 2021.

\bibitem{10.1145/3503927}
X.~Man, D.~Ouyang, X.~Li, J.~Song, and J.~Shao, ``Scenario-aware recurrent
  transformer for goal-directed video captioning,'' \emph{ACM Trans. Multimedia
  Comput. Commun. Appl.}, vol.~18, no.~4, 2022.

\bibitem{Yushchenko_2019_ICCV}
V.~Yushchenko, N.~Araslanov, and S.~Roth, ``Markov decision process for video
  generation,'' in \emph{Proceedings of the IEEE/CVF International Conference
  on Computer Vision Workshops}, 2019.

\bibitem{NIPS2016_04025959}
C.~Vondrick, H.~Pirsiavash, and A.~Torralba, ``Generating videos with scene
  dynamics,'' \emph{Advances in Neural Information Processing Systems},
  vol.~29, 2016.

\bibitem{Tulyakov_2018_CVPR}
S.~Tulyakov, M.-Y. Liu, X.~Yang, and J.~Kautz, ``Mocogan: Decomposing motion
  and content for video generation,'' in \emph{Proceedings of the IEEE
  Conference on Computer Vision and Pattern Recognition}, 2018, pp. 1526--1535.

\bibitem{TGAN}
M.~Saito, S.~Saito, M.~Koyama, and S.~Kobayashi, ``Train sparsely, generate
  densely: Memory-efficient unsupervised training of high-resolution temporal
  gan,'' \emph{International Journal of Computer Vision}, vol. 128, no.~10, pp.
  2586--2606, 2020.

\bibitem{NIPS2014_5ca3e9b1}
I.~Goodfellow, J.~Pouget-Abadie, M.~Mirza, B.~Xu, D.~Warde-Farley, S.~Ozair,
  A.~Courville, and Y.~Bengio, ``Generative adversarial nets,'' \emph{Advances
  in Neural Information Processing Systems}, vol.~27, 2014.

\bibitem{DBLP:conf/iclr/BrockDS19}
A.~Brock, J.~Donahue, and K.~Simonyan, ``Large scale {GAN} training for high
  fidelity natural image synthesis,'' in \emph{International Conference on
  Learning Representations}, 2019.

\bibitem{stylegan_v}
I.~Skorokhodov, S.~Tulyakov, and M.~Elhoseiny, ``Stylegan-v: A continuous video
  generator with the price, image quality and perks of stylegan2,'' in
  \emph{Proceedings of the IEEE/CVF Conference on Computer Vision and Pattern
  Recognition}, 2022, pp. 3626--3636.

\bibitem{2018faceforensics}
A.~Rössler, D.~Cozzolino, L.~Verdoliva, C.~Riess, J.~Thies, and M.~Nießner,
  ``Faceforensics: A large-scale video dataset for forgery detection in human
  faces,'' \emph{arXiv preprint arXiv:1803.09179}, 2018.

\bibitem{Xiong_2018_CVPR}
W.~Xiong, W.~Luo, L.~Ma, W.~Liu, and J.~Luo, ``Learning to generate time-lapse
  videos using multi-stage dynamic generative adversarial networks,'' in
  \emph{Proceedings of the IEEE Conference on Computer Vision and Pattern
  Recognition}, 2018, pp. 2364--2373.

\bibitem{tian2021a}
Y.~Tian, J.~Ren, M.~Chai, K.~Olszewski, X.~Peng, D.~N. Metaxas, and
  S.~Tulyakov, ``A good image generator is what you need for high-resolution
  video synthesis,'' in \emph{International Conference on Learning
  Representations}, 2021.

\bibitem{zhu2021lowrankgan}
J.~Zhu, R.~Feng, Y.~Shen, D.~Zhao, Z.-J. Zha, J.~Zhou, and Q.~Chen, ``Low-rank
  subspaces in gans,'' \emph{Advances in Neural Information Processing
  Systems}, vol.~34, pp. 16\,648--16\,658, 2021.

\bibitem{6795963}
S.~Hochreiter and J.~Schmidhuber, ``Long short-term memory,'' \emph{Neural
  Computation}, vol.~9, no.~8, pp. 1735--1780., 1997.

\bibitem{soomro2012ucf101}
K.~Soomro, A.~R. Zamir, and M.~Shah, ``Ucf101: A dataset of 101 human actions
  classes from videos in the wild,'' \emph{arXiv preprint arXiv:1212.0402},
  2012.

\bibitem{unterthiner2019accurate}
T.~Unterthiner, S.~van Steenkiste, K.~Kurach, R.~Marinier, M.~Michalski, and
  S.~Gelly, ``Towards accurate generative models of video: A new metric and
  challenges,'' \emph{arXiv preprint arXiv:1812.01717}, 2019.

\bibitem{Saito_2017_ICCV}
M.~Saito, E.~Matsumoto, and S.~Saito, ``Temporal generative adversarial nets
  with singular value clipping,'' in \emph{Proceedings of the IEEE
  International Conference on Computer Vision}, 2017, pp. 2830--2839.

\bibitem{RadfordMC15}
A.~Radford, L.~Metz, and S.~Chintala, ``Unsupervised representation learning
  with deep convolutional generative adversarial networks,'' in
  \emph{International Conference on Learning Representations}, 2016.

\bibitem{NIPS2017_d0010a6f}
I.~O. Tolstikhin, S.~Gelly, O.~Bousquet, C.-J. Simon-Gabriel, and
  B.~Sch{\"o}lkopf, ``Adagan: Boosting generative models,'' \emph{Advances in
  Neural Information Processing Systems}, vol.~30, 2017.

\bibitem{Gong_2019_ICCV}
X.~Gong, S.~Chang, Y.~Jiang, and Z.~Wang, ``Autogan: Neural architecture search
  for generative adversarial networks,'' in \emph{Proceedings of the IEEE/CVF
  International Conference on Computer Vision}, 2019, pp. 3224--3234.

\bibitem{2021you}
E.~Sch{\"o}nfeld, V.~Sushko, D.~Zhang, J.~Gall, B.~Schiele, and A.~Khoreva,
  ``You only need adversarial supervision for semantic image synthesis,'' in
  \emph{International Conference on Learning Representations}, 2021.

\bibitem{NIPS2016_8a3363ab}
T.~Salimans, I.~Goodfellow, W.~Zaremba, V.~Cheung, A.~Radford, and X.~Chen,
  ``Improved techniques for training gans,'' \emph{Advances in Neural
  Information Processing Systems}, vol.~29, 2016.

\bibitem{NIPS2017_892c3b1c}
I.~Gulrajani, F.~Ahmed, M.~Arjovsky, V.~Dumoulin, and A.~C. Courville,
  ``Improved training of wasserstein gans,'' \emph{Advances in Neural
  Information Processing Systems}, vol.~30, 2017.

\bibitem{10.1145/3343031.3350944}
H.~Wu, S.~Zheng, J.~Zhang, and K.~Huang, ``Gp-gan: Towards realistic
  high-resolution image blending,'' in \emph{Proceedings of the 27th ACM
  International Conference on Multimedia}, 2019, pp. 2487--2495.

\bibitem{NEURIPS2019_18cdf49e}
J.~Donahue and K.~Simonyan, ``Large scale adversarial representation
  learning,'' \emph{Advances in Neural Information Processing Systems},
  vol.~32, 2019.

\bibitem{DBLP:journals/corr/abs-1907-06571}
A.~Clark, J.~Donahue, and K.~Simonyan, ``Efficient video generation on complex
  datasets,'' \emph{arXiv preprint arXiv:1907.06571.}, 2019.

\bibitem{acharya2018high}
D.~Acharya, Z.~Huang, D.~P. Paudel, and L.~V. Gool, ``Towards high resolution
  video generation with progressive growing of sliced wasserstein gans,''
  \emph{arXiv preprint arXiv:1810.02419.}, 2018.

\bibitem{Aich_2020_CVPR}
A.~Aich, A.~Gupta, R.~Panda, R.~Hyder, M.~S. Asif, and A.~K. Roy-Chowdhury,
  ``Non-adversarial video synthesis with learned priors,'' in \emph{Proceedings
  of the IEEE/CVF Conference on Computer Vision and Pattern Recognition}, 2020,
  pp. 6090--6099.

\bibitem{Munoz_2021_WACV}
A.~Munoz, M.~Zolfaghari, M.~Argus, and T.~Brox, ``Temporal shift gan for large
  scale video generation,'' in \emph{Proceedings of the IEEE/CVF Winter
  Conference on Applications of Computer Vision}, 2021, pp. 3179--3188.

\bibitem{Sangeek_2021_CVPR}
S.~Hyun, J.~Kim, and J.-P. Heo, ``Self-supervised video gans: Learning for
  appearance consistency and motion coherency,'' in \emph{Proceedings of the
  IEEE/CVF Conference on Computer Vision and Pattern Recognition}, 2021, pp.
  10\,826--10\,835.

\bibitem{yan2021videogpt}
W.~Yan, Y.~Zhang, P.~Abbeel, and A.~Srinivas, ``Videogpt: Video generation
  using vq-vae and transformers,'' \emph{arXiv preprint arXiv:2104.10157},
  2021.

\bibitem{fox2021stylevideogan}
G.~Fox, A.~Tewari, M.~Elgharib, and C.~Theobalt, ``Stylevideogan: A temporal
  generative model using a pretrained stylegan,'' in \emph{British Machine
  Vision Conference}, 2021.

\bibitem{arjovsky2017wasserstein}
M.~Arjovsky, S.~Chintala, and L.~Bottou, ``{W}asserstein generative adversarial
  networks,'' in \emph{Proceedings of the 34th International Conference on
  Machine Learning}, vol.~70.\hskip 1em plus 0.5em minus 0.4em\relax PMLR,
  2017, pp. 214--223.

\bibitem{yu2022generating}
S.~Yu, J.~Tack, S.~Mo, H.~Kim, J.~Kim, J.-W. Ha, and J.~Shin, ``Generating
  videos with dynamics-aware implicit generative adversarial networks,'' in
  \emph{International Conference on Learning Representations}, 2022.

\bibitem{KAHEMBWE2020506}
E.~Kahembwe and S.~Ramamoorthy, ``Lower dimensional kernels for video
  discriminators,'' \emph{Neural Networks}, vol. 132, pp. 506--520, 2020.

\bibitem{DBLP:conf/iclr/BauZSZTFT19}
D.~Bau, J.~Zhu, H.~Strobelt, B.~Zhou, J.~B. Tenenbaum, W.~T. Freeman, and
  A.~Torralba, ``{GAN} dissection: Visualizing and understanding generative
  adversarial networks,'' in \emph{International Conference on Learning
  Representations}, 2019.

\bibitem{pmlr-v119-voynov20a}
A.~Voynov and A.~Babenko, ``Unsupervised discovery of interpretable directions
  in the gan latent space,'' in \emph{International Conference on Machine
  Learning}.\hskip 1em plus 0.5em minus 0.4em\relax PMLR, 2020, pp. 9786--9796.

\bibitem{Turkoglu_Thong_Spreeuwers_Kicanaoglu_2019}
M.~O. Turkoglu, W.~Thong, L.~Spreeuwers, and B.~Kicanaoglu, ``A layer-based
  sequential framework for scene generation with gans,'' in \emph{Proceedings
  of the AAAI Conference on Artificial Intelligence}, vol.~33, no.~01, 2019,
  pp. 8901--8908.

\bibitem{Goetschalckx_2019_ICCV}
L.~Goetschalckx, A.~Andonian, A.~Oliva, and P.~Isola, ``Ganalyze: Toward visual
  definitions of cognitive image properties,'' in \emph{Proceedings of the
  IEEE/CVF International Conference on Computer Vision}, 2019, pp. 5744--5753.

\bibitem{Shen_2020_CVPR}
Y.~Shen, J.~Gu, X.~Tang, and B.~Zhou, ``Interpreting the latent space of gans
  for semantic face editing,'' in \emph{Proceedings of the IEEE/CVF Conference
  on Computer Vision and Pattern Recognition}, 2020, pp. 9243--9252.

\bibitem{Jahanian*2020On}
A.~Jahanian, L.~Chai, and P.~Isola, ``On the "steerability" of generative
  adversarial networks,'' in \emph{International Conference on Learning
  Representations}, 2020.

\bibitem{StyleSpace}
Z.~Wu, D.~Lischinski, and E.~Shechtman, ``Stylespace analysis: Disentangled
  controls for stylegan image generation,'' in \emph{Proceedings of the
  IEEE/CVF Conference on Computer Vision and Pattern Recognition}, 2021, pp.
  12\,863--12\,872.

\bibitem{Abdal_2019_ICCV}
R.~Abdal, Y.~Qin, and P.~Wonka, ``Image2stylegan: How to embed images into the
  stylegan latent space?'' in \emph{Proceedings of the IEEE/CVF International
  Conference on Computer Vision}, 2019, pp. 4432--4441.

\bibitem{Abdal_2020_CVPR}
{R. Abdal, Y. Qin and P. Wonka}, ``Image2stylegan++: How to edit the embedded
  images?'' in \emph{Proceedings of the IEEE/CVF Conference on Computer Vision
  and Pattern Recognition}, 2020, pp. 8296--8305.

\bibitem{9241434}
Y.~Shen, C.~Yang, X.~Tang, and B.~Zhou, ``Interfacegan: Interpreting the
  disentangled face representation learned by gans,'' \emph{IEEE Transactions
  on Pattern Analysis and Machine Intelligence}, 2020.

\bibitem{Plumerault2020Controlling}
A.~Plumerault, H.~L. Borgne, and C.~Hudelot, ``Controlling generative models
  with continuous factors of variations,'' in \emph{International Conference on
  Learning Representations}, 2020.

\bibitem{yu2021}
Y.~Shen and B.~Zhou, ``Closed-form factorization of latent semantics in gans,''
  in \emph{Proceedings of the IEEE/CVF Conference on Computer Vision and
  Pattern Recognition}, 2021, pp. 1532--1540.

\bibitem{ling2021editgan}
H.~Ling, K.~Kreis, D.~Li, S.~W. Kim, A.~Torralba, and S.~Fidler, ``Editgan:
  High-precision semantic image editing,'' \emph{Advances in Neural Information
  Processing Systems}, vol.~34, pp. 16\,331--16\,345, 2021.

\bibitem{DatasetGAN}
Y.~Zhang, H.~Ling, J.~Gao, K.~Yin, J.-F. Lafleche, A.~Barriuso, A.~Torralba,
  and S.~Fidler, ``Datasetgan: Efficient labeled data factory with minimal
  human effort,'' in \emph{Proceedings of the IEEE/CVF Conference on Computer
  Vision and Pattern Recognition}, 2021, pp. 10\,145--10\,155.

\bibitem{SemanticGAN}
D.~Li, J.~Yang, K.~Kreis, A.~Torralba, and S.~Fidler, ``Semantic segmentation
  with generative models: Semi-supervised learning and strong out-of-domain
  generalization,'' in \emph{Proceedings of the IEEE/CVF Conference on Computer
  Vision and Pattern Recognition}, 2021, pp. 8300--8311.

\bibitem{10.1561/2200000016}
S.~Boyd, N.~Parikh, E.~Chu, B.~Peleato, J.~Eckstein \emph{et~al.},
  ``Distributed optimization and statistical learning via the alternating
  direction method of multipliers,'' \emph{Foundations and
  Trends{\textregistered} in Machine Learning}, vol.~3, no.~1, pp. 1--122,
  2011.

\bibitem{2013}
O.~Kuybeda, G.~A. Frank, A.~Bartesaghi, M.~Borgnia, S.~Subramaniam, and
  G.~Sapiro, ``A collaborative framework for 3d alignment and classification of
  heterogeneous subvolumes in cryo-electron tomography,'' \emph{Journal of
  Structural Biology}, vol. 181, no.~2, pp. 116--127, 2013.

\bibitem{10.5555/3295222.3295408}
M.~Heusel, H.~Ramsauer, T.~Unterthiner, B.~Nessler, and S.~Hochreiter, ``Gans
  trained by a two time-scale update rule converge to a local nash
  equilibrium,'' \emph{Advances in Neural Information Processing Systems},
  vol.~30, 2017.

\bibitem{kaisiyuan2020mead}
K.~Wang, Q.~Wu, L.~Song, Z.~Yang, W.~Wu, C.~Qian, R.~He, Y.~Qiao, and C.~C.
  Loy, ``Mead: A large-scale audio-visual dataset for emotional talking-face
  generation,'' in \emph{European Conference on Computer Vision}.\hskip 1em
  plus 0.5em minus 0.4em\relax Springer, 2020, pp. 700--717.

\bibitem{NEURIPS2020_8d30aa96}
T.~Karras, M.~Aittala, J.~Hellsten, S.~Laine, J.~Lehtinen, and T.~Aila,
  ``Training generative adversarial networks with limited data,''
  \emph{Advances in Neural Information Processing Systems}, vol.~33, pp.
  12\,104--12\,114, 2020.

\end{thebibliography}

\vspace{-33pt}
\begin{IEEEbiography}[{\includegraphics[width=1in,height=1.25in,clip,keepaspectratio]{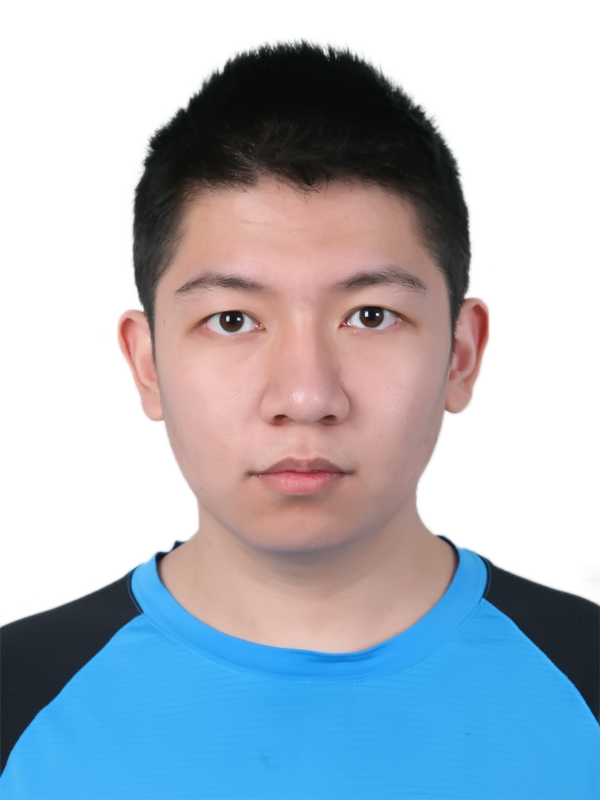}}]{Jingyuan Zhu} received the B.S. degree in Aerospace of Engineering of Tsinghua University in 2020, where he is currently working toward the Ph.D. degree in the Department of Electronic Engineering of Tsinghua University. 
\end{IEEEbiography}

\begin{IEEEbiography}[{\includegraphics[width=1in,height=1.25in,clip,keepaspectratio]{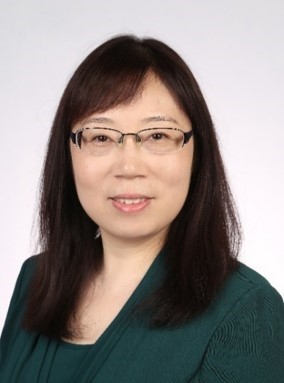}}]{Huimin Ma} received the M.S. and Ph.D. degrees in Mechanical Electronic Engineering from Beijing Institute of Technology, Beijing, China in 1998 and 2001, respectively. She is a professor in the School of Computer and Communication Engineering of University of Science and Technology Beijing. Her research and teaching interests include 3D object recognition and tracking, system modeling and simulation, psychological base of image cognition.
\end{IEEEbiography}

\begin{IEEEbiography}[{\includegraphics[width=1in,height=1.25in,clip,keepaspectratio]{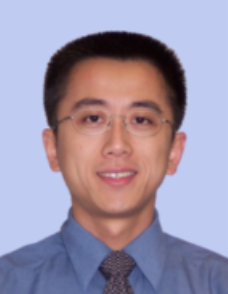}}]{Jiansheng Chen} received the M.S. degree in the Department of Computer Science and Technology from Tsinghua University and Ph.D. degree in the Computer Science and Technology from The Chinese University of Hong Kong in 2002 and 2006, respectively. He is a professor in the School of Computer and Communication Engineering of University of Science and Technology Beijing. His research and teaching interests include computer vision and machine learning.
\end{IEEEbiography}

\begin{IEEEbiography}[{\includegraphics[width=1in,height=1.25in,clip,keepaspectratio]{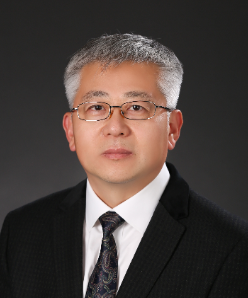}}]{Jian Yuan} received the Ph.D. degree in the Communication and Electronic Systems from University of Electronic Science and Technology of China. He is a professor in the Department of Electronic Engineering of Tsinghua University. His research and teaching interests include complex network theory and technologies.
\end{IEEEbiography}

\vfill

\end{document}